\documentclass{article}
\usepackage{iclr2027_conference,times}
\iclrfinalcopy

\usepackage[T1]{fontenc}
\usepackage[utf8]{inputenc}
\usepackage{microtype}

\usepackage{amsmath}
\usepackage{amssymb}
\usepackage{amsthm}
\usepackage{mathtools}
\usepackage{bm}

\usepackage{graphicx}
\usepackage{subcaption}
\usepackage{wrapfig}
\usepackage{float}

\usepackage{booktabs}
\usepackage{multirow}
\usepackage{array}
\usepackage{tabularx}
\usepackage{makecell}
\usepackage{adjustbox}  
\usepackage{siunitx}
\usepackage{colortbl}

\usepackage{xcolor}
\usepackage{url}
\usepackage[ruled,vlined]{algorithm2e}
\usepackage{enumitem}
\usepackage{xspace}
\usepackage{hyperref}  
\usepackage[capitalize,noabbrev]{cleveref}   

\hypersetup{colorlinks=true, linkcolor=black, citecolor=black, urlcolor=blue}

\usepackage{soul}   
\usepackage{placeins}
\usepackage{wrapfig}
\usepackage{needspace}

\theoremstyle{definition}

\title{Pruning DNA Sequence}

\author{Faias Satter\textsuperscript{1}, Shaker Islam\textsuperscript{2}, Aminul Islam\textsuperscript{1} \\
\textsuperscript{1}University of Louisiana at Lafayette \quad \textsuperscript{2}Johns Hopkins University \\
\texttt{faias.satter1@louisiana.edu}, \texttt{sislam30@jh.edu}, \texttt{aminul@louisiana.edu}}

\begin{document}

\maketitle
\lhead{Preprint}\vspace{-11.66pt}

\begin{abstract}
Data pruning can substantially reduce training cost by removing examples that contribute little additional learning signal, but it has not been studied for DNA language-model pretraining. We introduce Sequence Pruning (SP), a dynamic pruning framework that removes genomic sequences as they become easy for the model to predict. SP uses per-sequence pretraining accuracy computed directly from the model's forward pass, requiring no separate scoring model or additional pass over the corpus. We evaluate SP across five pretraining settings spanning two corpora, three model families, and both MLM and CLM objectives, with downstream evaluation on five genomic benchmarks. SP reduces pretraining data processing by up to 96\% while largely preserving downstream performance. At moderate savings, pruning improves performance by up to 1.2 MCC points over full-data pretraining; at the highest savings, pretraining uses only 4\% of the baseline GPU hours with little performance loss. SP also outperforms the dynamic pruning methods InfoBatch and BPAS across most comparisons. Beyond accuracy, our framework supports alternative criteria such as perplexity and EL2N, although accuracy yields the strongest downstream performance at matched savings. These results show that genomic pretraining does not require repeatedly processing all available sequences and that dynamic sequence-level pruning can substantially improve its data and computational efficiency.

\end{abstract}

\section{Introduction}
\label{sec:intro}

DNA, RNA, and proteins encode biological information as sequences over small alphabets, where the role of each element depends strongly on its surrounding context. DNA remains particularly difficult to interpret: only a small fraction of the human genome encodes proteins~\citep{lander2001human}, while the functions of much of the remaining sequence are still incompletely understood~\citep{encode2020expanded}. Functional labels therefore depend on experimental assays that are far more costly and slower to obtain than raw genomic sequence. As sequencing technologies continue to generate data at increasing scale, genomics faces a growing imbalance between abundant unlabeled sequence and scarce functional annotation. Self-supervised language modeling is well suited to this setting because it learns directly from raw sequence and transfers the resulting representations to downstream tasks. Genomic foundation models broadly follow two pretraining paradigms. Encoder-only models, including DNABERT (DB)~\citep{ji2021dnabert}, DNABERT-2 (DB2)~\citep{zhou2024dnabert2}, Nucleotide Transformer (NT)~\citep{dallatorre2025nucleotidetransformer}, and GENA-LM~\citep{10.1093/nar/gkae1310}, use masked language modeling (MLM), while decoder-only models such as HyenaDNA (Hyena)~\citep{hyenadna}, Evo~\citep{evo}, and Evo 2~\citep{evo2} use causal language modeling (CLM).

From a computational perspective, genomic foundation models have followed two broad scaling strategies. The first emphasizes efficiency, aiming to maintain downstream performance with fewer parameters and, in some cases, less pretraining data. For example, DB2, with 117M parameters, remains competitive with the 2.5B-parameter NT, while Hyena reports performance comparable to NT using \(1500\times\) fewer parameters and \(3200\times\) less pretraining data. The second strategy instead pursues performance through increasingly large models and pretraining corpora. Evo 2 scales to 40B parameters and is pretrained on 9.3T nucleotides, while NTv3~\citep{NTV3} is pretrained on 9T base pairs. Pretraining at this scale is inaccessible to most research groups, yet recent progress increasingly depends on larger computational budgets. At the same time, \citet{nopretrain} found that randomly initialized models can sometimes match or even outperform their pretrained counterparts on downstream tasks, suggesting that additional pretraining does not necessarily translate into better representations. Together, these observations raise a more fundamental question: How efficiently do genomic foundation models use their pretraining data to learn transferable representations?

Genomic pretraining datasets are themselves large. The Human Reference Genome (HRG) contains 2.75B bases~\citep{dallatorre2025nucleotidetransformer}, while the Multi-Species Genome collection (MSG) contains 32.49B~\citep{zhou2024dnabert2}. Because the computational cost of pretraining scales with the amount of sequence processed, larger corpora directly increase training cost. Yet it is unclear whether all sequences remain equally informative throughout pretraining. Some sequences may continue to provide useful learning signals, whereas others may become redundant once the model can predict them reliably. This motivates a more targeted question: Can a substantial fraction of DNA pretraining data be removed without degrading downstream performance?

Data pruning has been studied extensively in other domains. In vision, EL2N~\citep{el2n} scores each example by the \(\ell_2\) norm of its prediction error and removes low-scoring examples. \citet{marion2023moreinvestigatingdatapruning} evaluate the same criterion for LLM pretraining and compare it against perplexity-based pruning. Using perplexity, they show that training on 30\% of a web-text corpus can match full-data training. Both EL2N- and perplexity-based approaches are static: they score the corpus using a reference model before the pruned training run begins and perform pruning only once at initialization. Static pruning has also been studied for other biological sequences. \citet{wu2026investigatingdatapruningpretraining}, for example, prune RNA and protein corpora using a post-hoc influence criterion computed from a model already pretrained on the full corpus. Because this criterion itself requires a complete pretraining run, it is less suitable when the primary objective is to reduce pretraining cost. Dynamic pruning avoids this upfront requirement by updating sample importance during training. InfoBatch~\citep{infobatch} follows this strategy, while \citet{alam2026batch} outperform InfoBatch with batch pruning by activation stability (BPAS). We evaluate InfoBatch and BPAS as dynamic pruning baselines for DNA pretraining.

To the best of our knowledge, data pruning has not previously been studied for DNA language-model pretraining. We address this gap with Sequence Pruning (SP), a dynamic sequence-level pruning strategy based on a hypothesis: once a sequence is well learned, continuing to train on it provides diminishing benefit. Equivalently, easy sequences should contribute less useful learning signal than hard sequences. A similar pattern has been observed in web-text pretraining, where \citet{marion2023moreinvestigatingdatapruning} report that removing the easiest sequences can improve performance. This motivates a criterion for distinguishing easy from hard sequences during pretraining. Our approach is related to curriculum learning~\citep{curriculum}, which assigns examples a notion of difficulty, but differs in how that difficulty is used: rather than determining the order in which examples are presented, we use it to decide which sequences to remove. We define sequence difficulty using per-sequence pretraining accuracy (hereafter, \emph{accuracy}): MLM accuracy for encoder models and CLM accuracy for decoder models. Higher accuracy indicates that a sequence is easier and better learned; pruning such sequences therefore reduces computation spent on examples the model already predicts reliably.

\par
\Needspace{0\baselineskip}
\begin{wrapfigure}{r}{0.66\columnwidth}
\centering \vspace{-12pt}
\includegraphics[width=\linewidth]{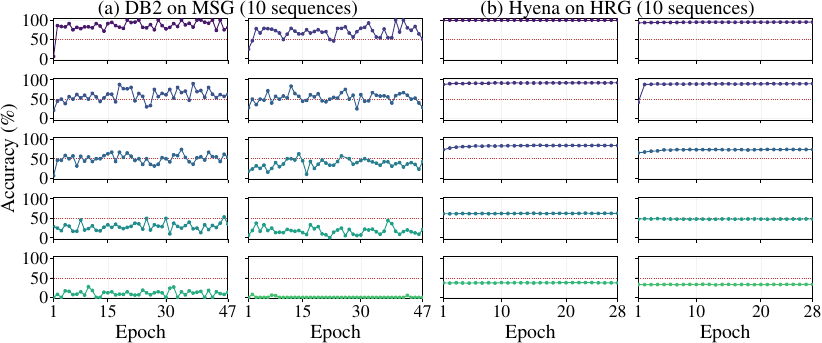} \vspace{-15pt}
\caption{Per-sequence pretraining accuracy trajectories for ten sequences each from DB2 (a) and Hyena (b). In each panel, the sequences are selected to span the observed accuracy range.} \vspace{-15pt}
\label{fig:acc_samples}
\end{wrapfigure}

\begin{wrapfigure}{r}{0.34\columnwidth}
\centering \vspace{-14pt}
\includegraphics[width=\linewidth]{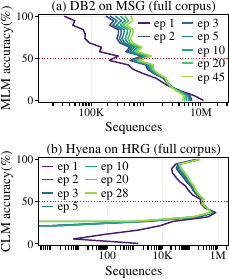} \vspace{-18pt}
\vspace{-1pt}\caption{Accuracy distributions across corpus at selected
epochs for DB2 (32.4M sequences, 47 epochs) and Hyena (2.7M sequences,
28 epochs). Curves show how sequence counts vary across the accuracy
range.} \vspace{-18pt}
\label{fig:acc_corpus}
\end{wrapfigure}

Can accuracy reliably distinguish easy sequences from hard ones? To investigate this, we track pretraining dynamics for DB2 and Hyena, covering both MLM and CLM objectives. Figure~\ref{fig:acc_samples} follows ten sequences from each setting and shows that per-sequence accuracy is largely stable across epochs: high-accuracy sequences tend to remain high, low-accuracy sequences remain low, and intermediate sequences stay in between. This stability does not mean that sequence accuracies remain unchanged: as the model learns, accuracies gradually shift upward while sequences largely remain within their relative difficulty bands. Figure~\ref{fig:acc_corpus} shows this behavior at the corpus level and makes the gradual improvement across epochs more visible. Each curve shows how many sequences fall into different accuracy ranges within one epoch, while curves from different epochs show how this distribution changes throughout pretraining. The first epoch has a distinct distribution because the model has only just begun learning the corpus. From the second epoch onward, however, the curves retain a similar overall shape while gradually shifting toward higher accuracy: fewer sequences remain in the low-accuracy ranges, while more move into the high-accuracy ranges. Thus, sequences identified as easy or hard early in pretraining tend to retain that status even as their accuracies improve. Beyond this coarse separation, accuracy also induces a finer-grained ordering from easiest to hardest, and after a single pass over the corpus, this ordering changes only marginally (Appendix~\ref{app:accuracy}). If accuracy is stable enough to preserve this finer ordering, it should also be sufficient for the coarser task of separating easy from hard sequences.

From which point should a sequence be considered easy or hard? This requires an accuracy threshold. Since easy sequences are the ones we aim to remove, we define the pruning range using two thresholds. The high threshold is fixed at perfect accuracy, while the low threshold, \(\tau_{\mathrm{low}}\), determines where pruning begins. We vary only \(\tau_{\mathrm{low}}\), making it the only tunable parameter introduced by our method. We therefore study how different values of \(\tau_{\mathrm{low}}\) affect downstream performance.

\par
\Needspace{10\baselineskip}

We study sequence pruning (SP) across five pretraining settings spanning two objectives, two corpora, and three model families, and evaluate downstream performance on five benchmarks. SP reduces pretraining data by 3\% to 96\%. At moderate savings, it not only preserves downstream performance but improves it by up to 1.2 MCC points over the unpruned baseline. At the highest savings, the model pretrains using only 4\% of the baseline GPU hours while maintaining comparable downstream performance. SP also outperforms the dynamic methods InfoBatch and BPAS: it achieves greater data savings and yields better downstream performance in most comparisons. We further incorporate two established pruning criteria, perplexity and EL2N, into our framework. Although both were originally designed for static pruning, recomputing them at each epoch boundary remains effective, demonstrating that our framework supports pruning criteria beyond accuracy. At matched savings, however, accuracy yields higher downstream performance than perplexity and EL2N. Our contributions are as follows.

1. We introduce data pruning to DNA pretraining. To the best of our knowledge, data pruning has not previously been applied to DNA corpora. We show that a large fraction of pretraining sequences can be removed without degrading downstream performance and, in several settings, that pruning can outperform full-data pretraining.

2. We introduce accuracy as a pruning criterion. Per-sequence pretraining accuracy consistently separates easy from hard sequences throughout pretraining and is obtained directly from the model's forward pass. It therefore requires no additional pass over the corpus and leaves the underlying pretraining recipe unchanged. Our method introduces only one tunable parameter, the low accuracy threshold, \(\tau_{\mathrm{low}}\). We identify two values that generalize across all settings we evaluate: one that yields the strongest downstream performance and one that achieves the largest data savings.

3. We propose a general plug-and-play framework for dynamic pruning of DNA sequences. The framework supports both MLM and CLM objectives and accommodates multiple pruning criteria, including accuracy, perplexity, and EL2N.

\begin{figure*}[htp]
\centering \vspace{-14pt}
\includegraphics[width=1.01\textwidth]{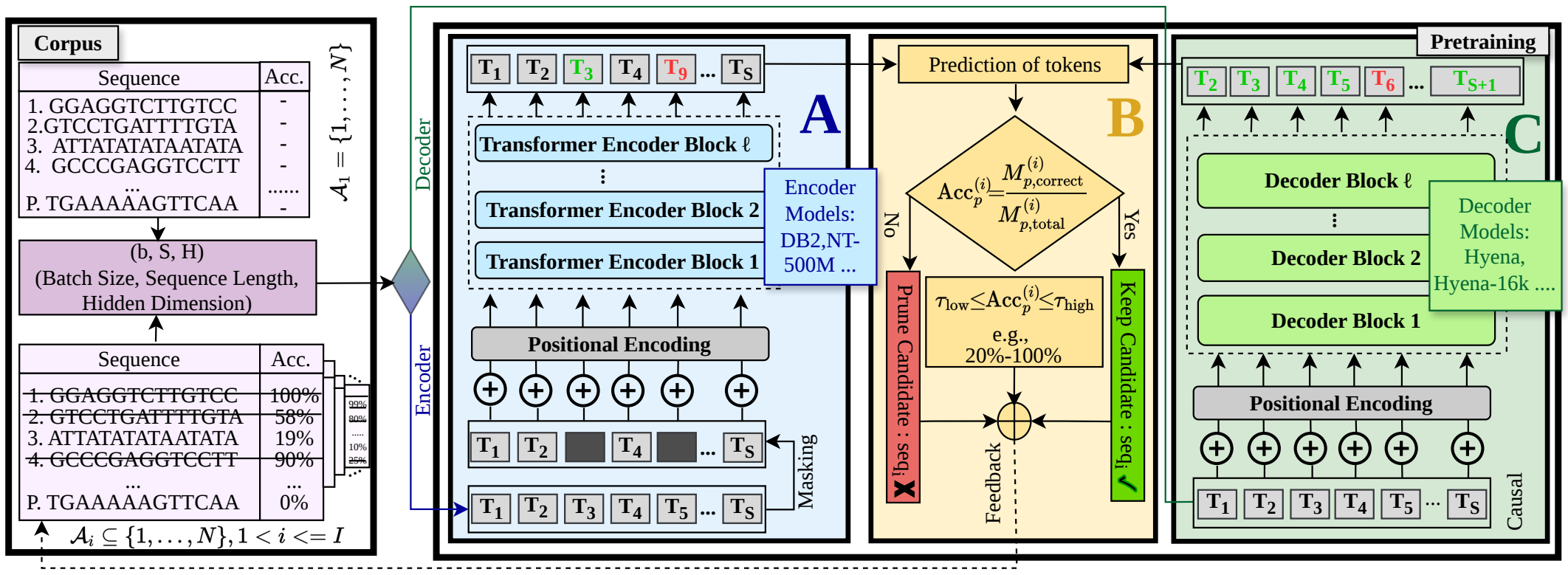} \vspace{-22pt}
\caption{Overview of the proposed SP framework. Left: the active corpus across epochs. Right: the pretraining pipeline, consisting of blocks A and B for MLM or blocks B and C for CLM, depending on the pretraining objective. Block B shows $[20\%,100\%]$ as an illustrative pruning range. T = Token.} \vspace{-16pt}
\label{fig:pipeline}
\end{figure*}

\section{Methodology}
\label{sec:method}

Figure~\ref{fig:pipeline} provides an overview of our pruning framework, which consists of two main components: the corpus and the pretraining model. Within the pretraining component, sequences pass through block A for MLM or block C for CLM, depending on the pretraining objective. At each epoch boundary, block B applies the pruning criterion and updates the corpus used in the following epoch. Thus, our method changes only which sequences are presented to the model across epochs. The model architecture, pretraining objective, optimizer, and downstream evaluation protocol remain unchanged from the corresponding unpruned setup.

A corpus contains \(N\) sequences indexed by \(p\), and a pretraining run consists of \(I\) epochs indexed by \(i\). At epoch \(i\), the model trains on an active corpus \(\mathcal{A}_i \subseteq \{1,\dots,N\}\), initialized with the full corpus, \(\mathcal{A}_1 = \{1,\dots,N\}\), corresponding to the corpus entering epoch 1 in Figure~\ref{fig:pipeline}. Sequences in \(\mathcal{A}_i\) are tokenized and grouped into batches, producing model inputs of shape \((b, S, H)\), where \(b\) is the batch size, \(S\) is the sequence length, and \(H\) is the hidden dimension. Each batch then follows block A or block C according to the model's pretraining objective. In the encoder branch (A), a fraction of tokens in each sequence is selected and masked. The sequence is passed through \(\ell\) transformer encoder blocks, and the model predicts the original tokens at the masked positions. In the decoder branch (C), the sequence passes through \(\ell\) causal decoder blocks, so each position can access only the tokens that precede it, and the model predicts each token from this context. Despite these different objectives, both branches produce token-level predictions at their respective prediction positions. We use the correctness of these predictions to compute per-sequence MLM accuracy for encoder models and per-sequence CLM accuracy for decoder models.

Let \(M_{p,\mathrm{total}}^{(i)}\) denote the number of prediction positions scored for sequence \(p\) at epoch \(i\): the masked positions for MLM and every position after the first for CLM. Let \(M_{p,\mathrm{correct}}^{(i)}\) denote the number of these positions predicted correctly. The per-sequence accuracy is \( \mathrm{Acc}_p^{(i)} \!=\! M_{p,\mathrm{correct}}^{(i)}/{M_{p,\mathrm{total}}^{(i)}}\), where \(M_{p,\mathrm{total}}^{(i)} \!>\! 0\). Because per-sequence accuracy is computed from the same token predictions used for the training loss, only \(M_{p,\mathrm{correct}}^{(i)}\) and \(M_{p,\mathrm{total}}^{(i)}\) need to be accumulated, with no additional forward pass. At the end of epoch \(i\), this yields one accuracy value for each scored sequence in \(\mathcal{A}_i\).

The accuracy definition is shared across both objectives, but the positions that contribute to it differ. Under MLM, the $80/10/10$ rule replaces selected tokens with \texttt{[MASK]}, random tokens, or leaves them unchanged. We score only the \texttt{[MASK]} positions to keep the accuracy calculation consistent across sequences. Because masking is resampled at every epoch, the accuracy of a sequence may be computed from different positions over time, introducing variability due to mask sampling. Despite this variability, the stability observed in Figure~\ref{fig:acc_samples} and~\ref{fig:acc_corpus} indicates that the distinction between easy and hard sequences is generally preserved. Under CLM, every position after the first contributes to the accuracy calculation, resulting in less epoch-to-epoch fluctuation than under MLM (Appendix~\ref{app:clm_mlm}).

At each epoch boundary, every sequence in $\mathcal{A}_i$ with valid accuracy is compared against the pruning thresholds,
\(\displaystyle
\tau_{\mathrm{low}} \;\le\; \mathrm{Acc}_p^{(i)} \;\le\; \tau_{\mathrm{high}},
\)
and sequences whose accuracy falls within this range are removed to form $\mathcal{A}_{i+1}$ for the next epoch. We fix $\tau_{\mathrm{high}} = 100\%$ for all SP variants, corresponding to perfect accuracy, making $\tau_{\mathrm{low}}$ the only parameter we vary. Lowering $\tau_{\mathrm{low}}$ expands the pruning range, making more sequences eligible for removal and pruning a larger fraction of the corpus. Since $\tau_{\mathrm{high}}$ remains fixed at $100\%$ accuracy, pruning is restricted to sequences the model already predicts well.

As the active corpus shrinks, each epoch of a pruned run requires fewer optimization steps than the unpruned baseline. Retaining the baseline step budget would therefore cause the remaining sequences to be processed more frequently, confounding data reduction with sample repetition. To avoid this, each pruned run retains the baseline epoch budget $I$ and terminates after at most $I$ epochs, or earlier if the active corpus becomes empty. This ensures that no retained sequence is processed more times than in the baseline. The learning-rate schedule remains tied to the baseline step count and is not rescaled, so the pruned run and baseline use the same learning rate at each matched optimization step. For InfoBatch and BPAS, block B in Figure~\ref{fig:pipeline} is replaced by each method's native pruning procedure, while the corresponding pretraining setting is otherwise kept fixed. For perplexity and EL2N, only the pruning criterion within block B is replaced, leaving the rest of the SP framework unchanged. Additional methodological details are provided in Appendix~\ref{app:framework_details}.

\section{Experiment}
\label{sec:setup}

\paragraph{Pretraining settings.}
A pretraining setting is defined by a specific combination of pretraining corpus and model. Table~\ref{tab:settings} summarizes the five settings, which vary in corpus, model scale, context length, and pretraining objective. Relative to DB2, NT-500M changes the model scale, Hyena changes the pretraining objective from MLM to CLM, and Hyena-16k increases the context length to $16{,}384$~bp. Each setting follows its published pretraining recipe (Appendix~\ref{app:pretraining_details}).

\par
\Needspace{10\baselineskip}
\begin{wrapfigure}[12]{r}{0.39\columnwidth}
\centering \vspace{-9pt}
\includegraphics[width=\linewidth]{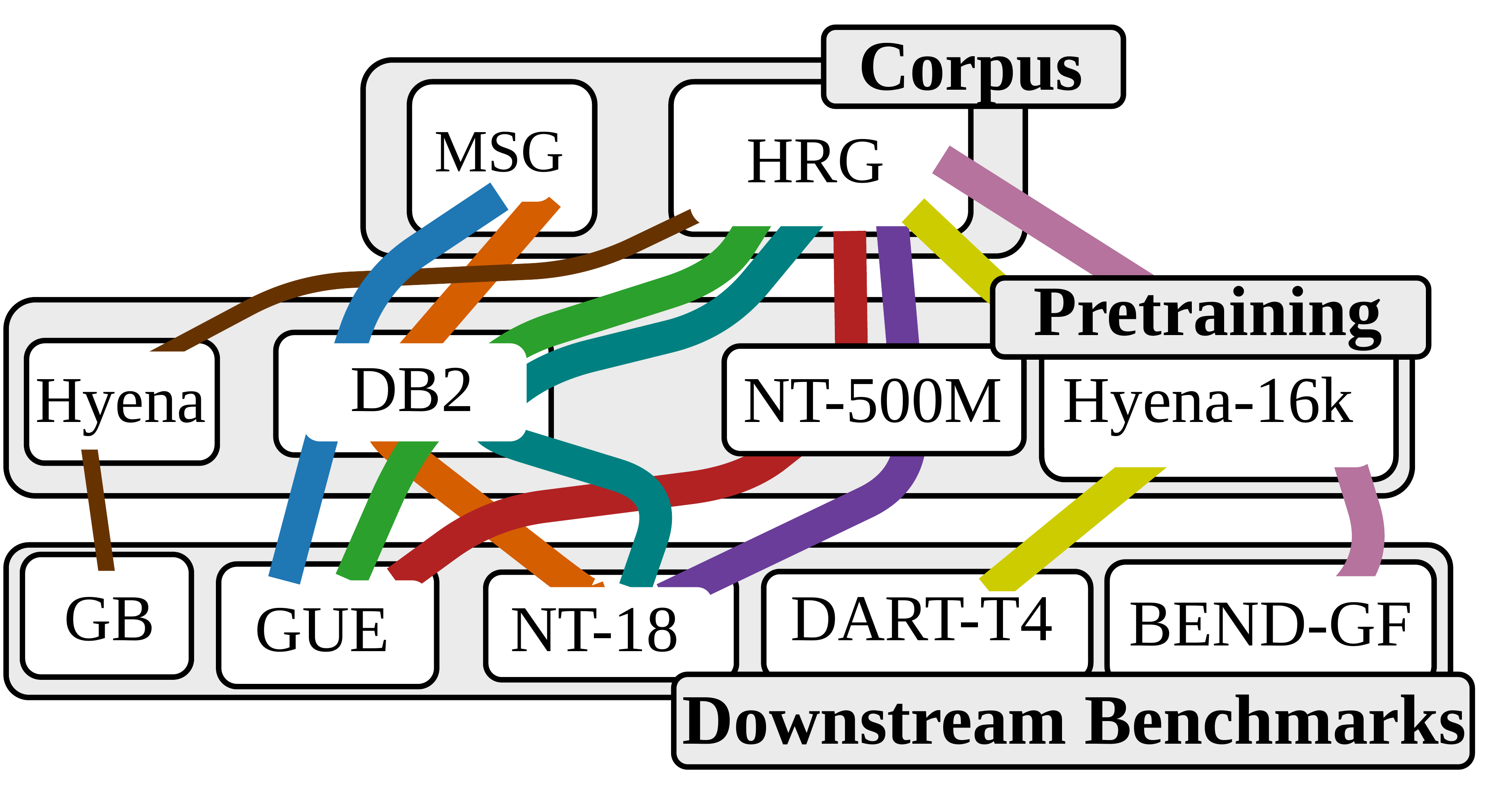} \vspace{-18pt}
\caption{Pretraining settings and corresponding downstream benchmarks; Edges indicating the evaluation paths.} \vspace{-14pt}
\label{fig:settings}
\end{wrapfigure}
\vspace{-5pt}

\vspace{-5pt}
\paragraph{Downstream benchmarks.}
Figure~\ref{fig:settings} summarizes the pretraining settings and their corresponding downstream evaluations. DB2 and NT-500M are evaluated on Genome Understanding Evaluation (GUE)~\citep{zhou2024dnabert2} and the Nucleotide Transformer benchmark (NT-18)~\citep{dallatorre2025nucleotidetransformer}, while Hyena is evaluated on Genomic Benchmarks (GB)~\citep{hyenadna}. These benchmarks use full-parameter fine-tuning, in which the entire pretrained model is updated for each task. Hyena-16k is evaluated on BEND~\citep{BEND} gene finding (B-GF) and DART-Eval~\citep{DART-eval} Task 4 (D-T4) using frozen probing, in which the pretrained model remains fixed and a small convolutional probe is trained on its embeddings. We report MCC for all benchmarks except D-T4, for which we report Spearman correlation. GUE and NT-18 organize their tasks into groups, so we first average performance within each group and then across groups, following their respective reporting protocols. For GB, we average performance directly across its seven tasks. Appendix~\ref{app:finetune} provides the full fine-tuning and frozen-probing protocols.

\vspace{-2pt}
\par
\Needspace{5\baselineskip}
\begin{wraptable}{R}{0.55\textwidth}
\centering
\small
\setlength{\tabcolsep}{1pt} \vspace{-15pt}
\caption{The five pretraining settings. BPE tokenizer denotes byte-pair encoding, and 6-mer denotes non-overlapping groups of six nucleotides.} \vspace{-6pt}
\label{tab:settings}
\begin{adjustbox}{max width=\linewidth}
\begin{tabular}{@{}llrlrrrr@{}}
\toprule
Corpus & Model & Param. & Toknzr. & Vocab & Seq. Len. & \# Seq. & Batch\\
\midrule
MSG & DB2       & 117M & BPE   & 4096 & 128     & 32.4M  & 3072\\
HRG & DB2       & 117M & BPE   & 4096 & 128     & 3M     & 3072\\
HRG & NT-500M   & 500M & 6-mer & 4105 & 1000    & 500k   & 1008\\
HRG & Hyena     & 6.6M & char  & 12   & 1024    & 2.7M   & 1536\\
HRG & Hyena-16k & 7.2M & char  & 12   & 16{,}384 & 185.6k & 96\\
\bottomrule \vspace{-18pt}
\end{tabular} 
\end{adjustbox}
\end{wraptable}

\paragraph{Variants and efficiency measures.}
\label{paragraph:eqs}

 Sequence Pruning (SP) variants are named by their $\tau_{\mathrm{low}}$. SP\_50\% sets $\tau_{\mathrm{low}}=50\%$ and $\tau_{\mathrm{high}}=100\%$, so at each epoch boundary, sequences with accuracy $50\%-100\%$ are pruned. For the other variants, the numerical suffixes only distinguish configurations; their parameters are listed in Appendix~\ref{app:hparams}. For efficiency measurement, DSI remains our primary measure. Following \citet{alam2026batch}, if $N$ is the number of sequences in the full corpus, $\mathcal{A}_i$ is the active corpus in epoch $i$, $I$ is the scheduled number of epochs, and $I_s$ is the stopping epoch, then $\mathrm{DSI}\!\!=\!\!1\!\!-\!\!\sum_{i=1}^{I_s}\!|\mathcal{A}_i|/(I\!\cdot\! N)$, where higher values indicate greater savings. $\mathrm{DSI}\!=\!50\%$ means that half of the baseline sequence processing is avoided. Additionally, we report GPU hours, computed as the number of GPUs multiplied by wall-clock training time. Pretraining uses 24 NVIDIA A100 GPUs, and fine-tuning uses one.

\subsection{Result Analysis}
\label{sec:results}

Table~\ref{tab:main_result_encoder} reports results for the three encoder pretraining settings: DB2 on MSG, DB2 on HRG, and NT-500M on HRG, while Table~\ref{tab:hyena-efficiency} reports Hyena on HRG, the decoder pretraining setting. Each setting includes two reference points: the unpruned baseline, which uses the full pretraining corpus throughout the run, and full pruned, which skips pretraining and uses the same fine-tuning protocol as our pruned variants. The downstream performance gap between these two reference points reflects the contribution of pretraining, which is the stage targeted by pruning. This gap is substantial for the encoder settings but smaller for Hyena. Hyena uses character-level tokenization, and \citet{nopretrain} similarly report limited gains from pretraining for character-tokenized models. For Hyena, SP\_50\% even outperforms the unpruned baseline. Across settings, SP remains substantially closer to the unpruned baseline than to full pruned, indicating that it preserves most of the benefit provided by pretraining.
Variant parameters and full pruning sweeps are provided in Appendix~\ref{app:sweeps}.

\par
\Needspace{40\baselineskip}
\begin{wraptable}{R}{0.53\textwidth}
\centering \vspace{-4pt}
\caption{Efficiency and downstream performance across the three encoder pretraining settings.} \vspace{-6pt}
\label{tab:main_result_encoder}
\renewcommand{\tabcolsep}{1pt}
\renewcommand{\arraystretch}{0.9}

\begin{adjustbox}{
    max totalsize={\linewidth}{0.82\textheight},
    center
}
\begin{tabular}{l@{\hspace{4pt}}llcccccc}
\toprule
 & Pruning
 & \multirow{2}{*}{Variant}
 & DSI
 & GPU$\downarrow$
 & \multirow{2}{*}{\rotatebox[origin=c]{90}{Epch.}}
 & \multirow{2}{*}{\rotatebox[origin=c]{90}{\shortstack{Steps\\(k)}}}
 & \multicolumn{2}{c}{MCC$\uparrow$} \\
\cmidrule(lr){8-9}
 & Apprch. & & \%$\uparrow$ & Hours & & & GUE & NT-18 \\
\midrule

\multirow{14}{*}{\rotatebox[origin=c]{90}{DB2 on MSG}}
 & \multirow{2}{*}{Baseline} & Unprun. & 0 & 1012 & 47 & 500 & \textbf{71.6} & \textbf{69.8} \\
 & & Full prun. & 100 & 0 & 0 & 0 & 60.6 & 47.1 \\
\cmidrule(lr){2-9}

 & \multirow{5}{*}{Related} & BPAS\_1 & 6 & 977 & 47 & 470 & 71.3 & 68.8 \\
 & \multirow{5}{*}{Work} & BPAS\_2 & 19 & 777 & 47 & 400 & 71.2 & \underline{69.6} \\
 & & BPAS\_4 & 36 & 838 & 38 & 310 & 71.0 & 68.6 \\
 & & BPAS\_9 & 71 & 318 & 24 & 140 & 68.9 & 67.5 \\
\cmidrule(lr){3-9}
 & & Info\_1 & 19 & 709 & 47 & 400 & 69.8 & 68.6 \\
 & & Info\_2 & 29 & 600 & 47 & 349 & 39.4 & 48.3 \\

\cmidrule(lr){2-9}
 & \multirow{4}{*}{Ours} & SP\_100\% & 5 & 994 & 47 & 480 & \underline{71.5} & 69.5 \\
 & & SP\_80\% & 12 & 905 & 47 & 430 & 70.8 & 69.2 \\
 & & SP\_50\% & 32 & 713 & 47 & 330 & 71.5 & 69.0 \\
 & & SP\_20\% & 77 & 226 & 47 & 110 & 70.5 & 69.1 \\
\cmidrule(lr){2-9}

 & \multirow{2}{*}{Ablation} & RP\_20\% & 77 & 215 & 47 & 110 & 68.1 & 65.6 \\
 & & LP\_0\% & 55 & 378 & 47 & 230 & 68.5 & 64.9 \\

\midrule

\multirow{13}{*}{\rotatebox[origin=c]{90}{DB2 on HRG}}
 & \multirow{2}{*}{Baseline} & Unprun. & 0 & 47 & 32 & 30 & \textbf{66.6} & 64.1 \\
 & & Full prun. & 100 & 0 & 0 & 0 & 43.6 & 47.3 \\
\cmidrule(lr){2-9}

 & \multirow{2}{*}{Related} & BPAS\_1 & 0 & 47 & 32 & 30 & 65.8 & 62.7 \\
 & \multirow{2}{*}{Work} & BPAS\_4 & 29 & 32 & 31 & 21 & 65.0 & 62.4 \\
 & & BPAS\_5 & 40 & 23 & 29 & 18 & 64.4 & 62.8 \\

\cmidrule(lr){2-9}
 & \multirow{4}{*}{Ours} & SP\_100\% & 2 & 46 & 32 & 29 & 65.2 & 63.8 \\
 & & SP\_80\% & 5 & 45 & 32 & 28 & 65.4 & \textbf{65.7} \\
 & & SP\_50\% & 18 & 39 & 32 & 25 & \underline{66.1} & 63.6 \\
 & & SP\_20\% & 46 & 25 & 32 & 16 & 65.2 & 63.2 \\
\cmidrule(lr){2-9}

 & \multirow{4}{*}{Ablation} & EL2N\_1 & 15 & 39 & 32 & 25 & 64.2 & \underline{64.7} \\
 & & EL2N\_2 & 25 & 34 & 32 & 22 & 65.3 & 64.6 \\
 & & EL2N\_3 & 37 & 32 & 32 & 19 & 65.8 & 64.2 \\
 & & EL2N\_4 & 44 & 27 & 32 & 16 & 65.0 & 63.0 \\

\midrule

\multirow{9}{*}{\rotatebox[origin=c]{90}{NT-500M on HRG}}
 & \multirow{2}{*}{Baseline} & Unprun. & 0 & 1137 & 99 & 50 & 67.0 & \textbf{74.2} \\
 & & Full prun. & 100 & 0 & 0 & 0 & 53.0 & 63.7 \\
\cmidrule(lr){2-9}

 & \multirow{2}{*}{Related} & BPAS\_1 & 1 & 1087 & 99 & 49 & 66.4 & 73.8 \\
 & \multirow{2}{*}{Work} & BPAS\_2 & 7 & 1033 & 99 & 46 & 65.6 & 73.7 \\
 & & BPAS\_4 & 40 & 658 & 99 & 29 & 65.6 & 73.9 \\

\cmidrule(lr){2-9}
 & \multirow{4}{*}{Ours} & SP\_100\% & 4 & 1081 & 99 & 47 & 67.4 & 73.9 \\
 & & SP\_80\% & 6 & 1063 & 99 & 46 & 67.1 & 73.9 \\
 & & SP\_50\% & 10 & 969 & 99 & 44 & \textbf{68.2} & 74.0 \\
 & & SP\_20\% & 37 & 668 & 99 & 30 & \underline{67.4} & \underline{74.1} \\

\bottomrule \vspace{-26pt}
\end{tabular} 
\end{adjustbox}
\end{wraptable}

Removing easy, well-learned sequences is more effective than removing random or hard sequences. We verify this with two control experiments. The first is random pruning (RP). For RP\_20\%, we match the DSI of SP\_20\% by randomly removing the same fraction of the corpus before pretraining and training for the same number of epochs. At matched DSI, RP\_20\% performs worse than SP\_20\% in both encoder and decoder settings, showing that data reduction alone does not reproduce the gains from SP. The second control is low-accuracy pruning (LP). Rather than removing easy sequences, LP\_0\% removes hard sequences at every epoch boundary by setting $\tau_{\mathrm{low}}=\tau_{\mathrm{high}}=0\%$. Thus, only sequences with 0\% accuracy are removed. This is the only variant in the paper for which the high pruning threshold ($\tau_{\mathrm{high}}$) differs from 100\%. On DB2--MSG, LP\_0\% achieves performance similar to RP\_20\%.

\vspace{-3pt}
\paragraph{Comparison with dynamic pruning.}
\label{subsec:dynamic}

In Table~\ref{tab:main_result_encoder}, Info\_1 uses the pruning ratio and annealing point recommended by \citet{infobatch} and reaches about 19\% DSI. Info\_2 targets higher DSI but performs below the full pruned DB2 baseline, suggesting that InfoBatch may require modifications beyond its recommended recipe for DNA pretraining. BPAS degrades more gradually and provides the stronger dynamic-pruning comparison. Table~\ref{tab:main_result_encoder} and Table~\ref{tab:hyena-efficiency} compare BPAS with SP at similar DSI. On DB2--MSG, SP remains competitive with the unpruned baseline at about 77\% DSI, whereas BPAS degrades sharply by 71\% DSI. On DB2--HRG, SP\_20\% achieves greater savings while attaining a higher downstream score than BPAS. On NT-500M, SP performs better at every compared DSI on both benchmarks, while SP\_50\% remains above the unpruned baseline on GUE. On Hyena, SP matches BPAS at the highest savings while reaching 96\% DSI versus 92\%.

\vspace{-5pt}
\paragraph{Other pruning criteria.}
\label{subsec:signals}
For a fair comparison, we integrate perplexity and EL2N into our framework and recompute them at every epoch boundary. For CLM, we follow \citet{marion2023moreinvestigatingdatapruning} to compute per-sequence perplexity and evaluate it on Hyena. For MLM, we evaluate per-sequence EL2N on DB2--HRG following \citet{el2n}. For both criteria, we use four thresholds chosen to match the DSI of four SP variants in Table~\ref{tab:main_result_encoder} and Table~\ref{tab:hyena-efficiency}; the corresponding thresholds are listed in Appendix~\ref{app:hparams}. Table~\ref{tab:hyena-efficiency} shows that Perp\_4 performs worse than SP and BPAS at comparable DSI. At matched DSI, EL2N yields slightly lower downstream performance than SP at nearly every comparison point. Downstream results show that the framework is not limited to accuracy, as both perplexity and EL2N can also serve as pruning criteria. Among the three criteria, accuracy achieves the strongest downstream performance, with a small but consistent advantage over the other two. Accuracy thresholds are also directly interpretable: a threshold of 20\% or 50\% specifies the per-sequence pretraining accuracy above which sequences become eligible for pruning. In contrast, InfoBatch relies on loss-based selection, BPAS on activation differences, perplexity on likelihood-based scores, and EL2N on error norms. These quantities are less directly interpretable and generally require tuning across settings.

\par
\Needspace{20\baselineskip}
\begin{wraptable}{R}{0.48\textwidth}
\renewcommand{\tabcolsep}{1pt}
\centering \vspace{-5pt}
\caption{Efficiency and downstream performance for decoder pretraining setting (Hyena on HRG).} \vspace{-5pt}
\label{tab:hyena-efficiency}
\begin{tabular}{lllccccc}
\toprule
 & Pruning
 & \multirow{2}{*}{Variant}
 & DSI
 & GPU $\downarrow$
 & \multirow{2}{*}{\rotatebox[origin=c]{90}{Epch.}}
 & \multirow{2}{*}{\rotatebox[origin=c]{90}{\shortstack{Steps\\(k)}}}
 & MCC$\uparrow$ \\
\cmidrule(lr){8-8}
 & Apprch. & & \%$\uparrow$ & Hours & & & GB \\
\midrule

 & \multirow{2}{*}{Baseline} & Unprun. & 0 & 76 & 28 & 50 & 77.0 \\
 & & Full prun. & 100 & 0 & 0 & 0 & 73.3 \\
\cmidrule(lr){1-8}
 & \multirow{3}{*}{Related} & BPAS\_4 & 2 & 73 & 28 & 48 & 77.2 \\
 & \multirow{3}{*}{Work} & BPAS\_6 & 9 & 68 & 28 & 45 & \textbf{77.3} \\
 & & BPAS\_9 & 63 & 27 & 23 & 18 & \underline{77.2} \\
 & & BPAS\_12 & 92 & 4 & 2 & 4 & 76.1\\
\cmidrule(lr){1-8}
 & \multirow{4}{*}{Ours} & SP\_100\% & 4 & 71 & 28 & 47 & 76.4 \\
 & & SP\_80\% & 11 & 65 & 28 & 43 & 76.4 \\
 & & SP\_50\% & 47 & 39 & 28 & 26 & \textbf{77.3} \\
 & & SP\_20\% & 96 & 2 & 28 & 1.7 & 76.1 \\
\cmidrule(lr){1-8}
 & \multirow{5}{*}{Ablation} & Perp\_1 & 6 & 71 & 28 & 47 & 75.9 \\
 & & Perp\_2 & 20 & 51 & 28 & 39 & 76.5 \\
 & & Perp\_3 & 45 & 36 & 28 & 27 & 76.7 \\

 & & Perp\_4 & 96 & 2 & 28 & 1.7 & 75.4 \\
\cmidrule(lr){3-8}
 & & RP\_20\% & 96 & 3 & 28 & 1.7 & 74.8 \\

\bottomrule
\end{tabular} \vspace{-12pt}
\end{wraptable}

\paragraph{Robustness on harder benchmarks.}
\label{subsec:harder}

To test robustness, we add two benchmarks that differ from the others in important ways: B-GF requires a prediction at every nucleotide over sequences up to $14{,}000$~bp, while D-T4 is a regression task evaluated by correlation rather than classification. To accommodate the longer B-GF sequences, we pretrain Hyena-16k at $16{,}384$~bp and evaluate it on both benchmarks. Hyena-16k is pretrained for only two epochs, allowing a single pruning decision after the first epoch. Table~\ref{tab:harder-benchmark} shows that all SP variants remain close to the unpruned baseline on both tasks, indicating that SP preserves the information required for these tasks despite having only one opportunity to prune. This result is consistent with Figure~\ref{fig:acc_corpus}, where the separation between easy and hard sequences emerges early in pretraining.

\vspace{-5pt}

\paragraph{Are these differences real?}
\label{subsec:seeds}

We measure run-to-run variation at both the fine-tuning and pretraining stages. Across repeated runs, the largest MCC standard deviation is 0.6 points, while SP improves over the unpruned baseline by up to 1.2 MCC points in Table~\ref{tab:main_result_encoder} and~\ref{tab:hyena-efficiency}. Data savings are also stable, with DSI standard deviations of at most 0.15 percentage points across repeated SP\_20\% pretraining runs. Together, these results show that both downstream performance and data savings are consistent across runs, and that the largest observed performance gain is larger than the measured run-to-run variation (Appendix~\ref{app:variance}).

\par
\Needspace{0\baselineskip}
\begin{wraptable}{r}{0.54\textwidth}
\renewcommand{\tabcolsep}{1pt}
\centering \vspace{-24pt}
\caption{Results on harder downstream benchmarks under frozen probing (Hyena-16k  on HRG).} \vspace{-7pt}
\label{tab:harder-benchmark}
\begin{tabular}{llcccccc}
\toprule
Pruning
 & \multirow{2}{*}{Variant}
 & DSI
 & GPU$\downarrow$
 & \multirow{2}{*}{\rotatebox[origin=c]{90}{Epch.}}
 & \multirow{2}{*}{\rotatebox[origin=c]{90}{\shortstack{Steps\\(k)}}}
 & MCC$\uparrow$
 & Sprmn$\uparrow$ \\
\cmidrule(lr){7-7}\cmidrule(lr){8-8}
Apprch. & & \%$\uparrow$ & Hours & & & B-GF & D-T4 \\
\midrule

 Baseline & Unprun. & 0 & 7 & 2 & 4 & \textbf{35.0} & 36.6 \\

\cmidrule(lr){1-8}
 \multirow{5}{*}{Ours} & SP\_100\% & 2 & 7 & 2 & 4 & 34.5 & \textbf{38.2} \\
 & SP\_80\% & 3 & 7 & 2 & 4 & 34.4 & 34.6 \\
 & SP\_50\% & 10 & 6 & 2 & 3.5 & \underline{34.9} & 34.6 \\
 & SP\_20\% & 49 & 4 & 2 & 2 & 33.0 & 37.7 \\
 & SP\_10\% & 49 & 4
 & 2 & 2 & 32.6 & \underline{37.8} \\

\bottomrule
\end{tabular} \vspace{-8pt}
\end{wraptable}

\vspace{-5pt}
\paragraph{How far pruning can go depends on the pretraining setting.}
\label{subsec:reach}

The savings achieved by a given $\tau_{\mathrm{low}}$ are not a property of the pruning method alone. The two DB2 settings in Table~\ref{tab:main_result_encoder} use the same architecture and differ only in corpus, yet the same parameter yields substantially different savings. This allows us to isolate the effect of the corpus on data savings while holding the model architecture fixed. Conversely, the three HRG settings in Table~\ref{tab:main_result_encoder} and~\ref{tab:hyena-efficiency} share the same corpus but use different models, and their savings also vary widely. The corpus and model therefore jointly determine how much of the data becomes well learned enough to prune, and how quickly this occurs. What remains consistent

\par
\Needspace{10\baselineskip}
\begin{wraptable}{R}{0.52\textwidth}
\renewcommand{\tabcolsep}{1pt}
\centering \vspace{-3pt}
\caption{Effect of averaging-window size on efficiency and downstream performance (DB2 on MSG).} \vspace{-7pt}
\label{tab:average x window}
\begin{tabular}{llccccccc}
\toprule
Pruning
 & \multirow{2}{*}{Variant}
 & \multirow{2}{*}{\rotatebox[origin=c]{90}{Wndw.}}
 & DSI
 & GPU$\downarrow$
 & \multirow{2}{*}{\rotatebox[origin=c]{90}{Epch.}}
 & \multirow{2}{*}{\rotatebox[origin=c]{90}{\shortstack{Steps\\(k)}}}
 & \multicolumn{2}{c}{MCC$\uparrow$} \\
\cmidrule(lr){8-9}
Apprch. & & & \%$\uparrow$ & Hours & & & GUE & NT-18 \\
\midrule
Baseline & Unprun. & - & 0 & 1012 & 47 & 500 & \underline{71.6} & \textbf{69.8} \\
\cmidrule(lr){1-9}
Ours & SP\_20\% & 1 & \textbf{77} & 226 & 47 & 110 & 70.5 & 69.1 \\
\cmidrule(lr){1-9}
\multirow{4}{*}{Ablation} & SP\_20\% & 2 & \underline{63} & 330 & 47 & 180 & 70.5 & \underline{69.4} \\
 & SP\_20\% & 3 & 56 & 359 & 47 & 210 & \textbf{71.7} & \textbf{69.8} \\
 & SP\_20\% & 4 & 51 & 433 & 47 & 237 & 71.3 & \textbf{69.8} \\
 & SP\_20\% & 5 & 48 & 479 & 47 & 253 & 70.6 & 69.2 \\
\bottomrule
\end{tabular} \vspace{-8pt}
\end{wraptable}
\vspace{-6pt}

across settings is the useful range of the parameter. SP\_50\% gives the strongest downstream performance across settings, while SP\_20\% is the most aggressive setting before performance begins to degrade meaningfully, although the savings achieved at these settings differ across pretraining configurations.

\vspace{-5pt}

\paragraph{Effect of averaging window.}
\label{subsec:ablation-window}

Instead of pruning based on per-sequence pretraining accuracy from a single epoch, we average each sequence's accuracy over a window of consecutive epochs before applying the pruning criterion. Table~\ref{tab:average x window} shows the trade-off for SP\_20\% on DB2--MSG. The one-window setting is identical to SP\_20\% in Table~\ref{tab:main_result_encoder}. Increasing the window stabilizes the pruning decision but reduces DSI. A window of three provides the clearest trade-off: DSI decreases to 56\%, while performance on both benchmarks exceeds that of the unpruned DB2 baseline. Thus, windowed accuracy provides an operating mode when stronger downstream performance is preferred over maximizing data savings.

\vspace{-5pt}

\par
\Needspace{10\baselineskip}
\begin{wrapfigure}{R}{0.40\textwidth}
\centering \vspace{-18pt}
\includegraphics[width=\linewidth]{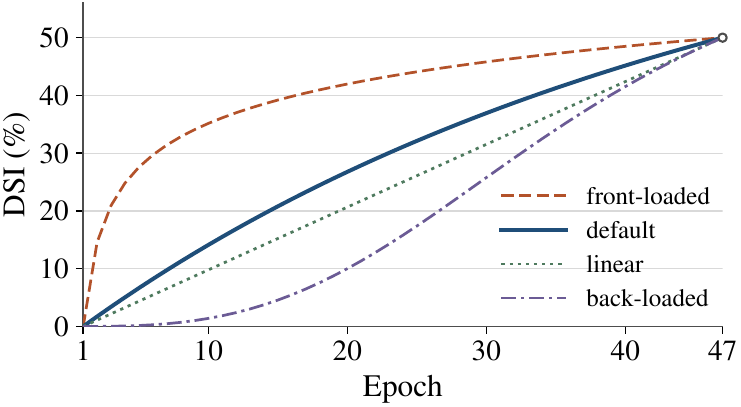}\vspace{-8pt}
\caption{Four pruning budgets that all reach $\mathrm{DSI} = 50\%$ at epoch 47 by different routes. The default is the trajectory measured in our SP runs on DB2--MSG.} \vspace{-10pt}
\label{fig:dsi-schedules}
\end{wrapfigure}

\vspace{-5pt}
\paragraph{DSI as control parameter.}
\label{subsec:ablation-dsi}

For a fixed $\tau_{\mathrm{low}}$, the achieved DSI is known only after the run. Across our variants, however, DSI follows a similar trajectory: it rises rapidly early in pretraining and then gradually flattens. We use this empirical pattern to define a default target DSI curve, turning DSI from an observed outcome into a controllable parameter (Figure~\ref{fig:dsi-schedules}). The curve specifies the target DSI after each epoch and, consequently, how many sequences may remain active. At each epoch boundary, we rank the remaining sequences by their per-sequence pretraining accuracy and remove the highest-scoring sequences until the target DSI is reached. Thus, the DSI curve determines how much to prune, while accuracy determines which sequences to remove.

\par
\Needspace{12\baselineskip}
\begin{wraptable}{R}{0.64\textwidth}
\renewcommand{\tabcolsep}{1pt}
\centering \vspace{-12pt}
\caption{Effect of the DSI target and schedule on efficiency and downstream performance (DB2 on MSG).}\vspace{-5pt}
\label{tab:bpas-efficiency}
\begin{tabular}{llccccccc}
\toprule
Pruning
 & \multirow{2}{*}{Variant}
 & DSI
 & DSI
 & GPU$\downarrow$
 & \multirow{2}{*}{\rotatebox[origin=c]{90}{Epch.}}
 & \multirow{2}{*}{\rotatebox[origin=c]{90}{\shortstack{Steps\\(k)}}}
 & \multicolumn{2}{c}{MCC$\uparrow$} \\
\cmidrule(lr){8-9}
Apprch. & & Schedule & \%$\uparrow$ & Hours & & & GUE & NT-18 \\
\midrule
Baseline & Unprun. & - & 0 & 1012 & 47 & 500 & 71.6 & \underline{69.8} \\
\cmidrule(lr){1-9}

\multirow{6}{*}{Ablation} & DSI\_20\% & default & 20 & 699 & 47 & 396 & \underline{71.8} & 69.4 \\
\cmidrule(lr){2-9}

& \multirow{4}{*}{DSI\_50\%} & default & 50 & 450 & 47 & 247 & \textbf{72.0} & \textbf{70.4} \\
& & front-loaded & 50 & 418 & 47 & 247 & 71.3 & 69.0 \\
& & back-loaded & 50 & 448 & 47 & 247 & 69.7 & 69.7 \\
& & linear & 50 & 452 & 46 & 247 & 70.2 & 68.4 \\
\cmidrule(lr){2-9}

& DSI\_80\% & default & 80 & 166 & 47 & 98 & 69.6 & 69.3 \\

\bottomrule 
\end{tabular} \vspace{-12pt}
\end{wraptable}

Table~\ref{tab:bpas-efficiency} shows that the default curve reproduces the SP results at comparable savings. To test whether DSI alone determines performance, we compare the default curve with three alternative curves that reach the same final DSI but distribute pruning differently across epochs. At $\mathrm{DSI}=50\%$, the default curve outperforms the unpruned baseline on both tasks while avoiding half of the baseline sequence processing. All three alternative schedules perform worse, showing that when sequences are removed matters in addition to how many are removed. Appendix~\ref{app:dsi} provides details.

\par
\Needspace{0\baselineskip}
\begin{wrapfigure}{R}{0.71\columnwidth}
    \centering \vspace{-28pt}
    \includegraphics[width=\linewidth]{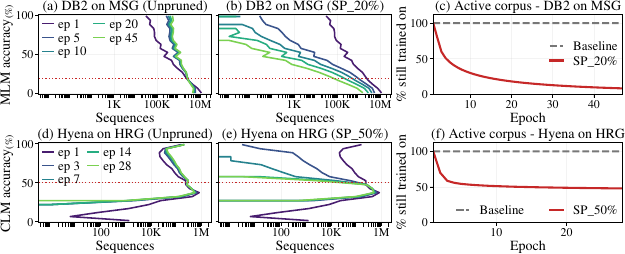}
    \vspace{-13pt}\caption{What pruning removes, and when. Panels (a) and (d) show how
sequence counts vary across the accuracy range when nothing is pruned,
and panels (b) and (e) show the same under sequence pruning. Panels (c)
and (f) show how much of the corpus is still being pretrained on at each
epoch. Dotted lines mark the $\tau_{\mathrm{low}}$.} \vspace{-18pt}
    \label{fig:pruning_dynamics}
\end{wrapfigure}

\vspace{-5pt}
\subsection{Why SP Works}
\label{sec:why}

Figure~\ref{fig:acc_corpus} shows that accuracy separates easy and hard sequences. Figure~\ref{fig:pruning_dynamics} then examines how pruning changes this separation in the same settings, comparing the DB2--MSG unpruned baseline with SP\_20\% and the Hyena--HRG unpruned baseline with SP\_50\%.

\vspace{-5pt}
\paragraph{The model's continued learning makes pruning dynamic.}
Although pruning removes high-accuracy sequences, prunable sequences do not run out after the first few epochs. Figure~\ref{fig:pruning_dynamics}(c) and (f) show that the active corpus continues to shrink throughout pretraining. On DB2, epoch 1 removes 3.24\% of the corpus, epoch 46 still removes 0.24\%, and roughly 40\% of all pruning occurs after epoch 10. This continued pruning arises because per-sequence accuracy gradually increases as the model learns to predict sequences better. As additional sequences cross $\tau_{\mathrm{low}}$, they become eligible for pruning in later epochs. Thus, although $\tau_{\mathrm{low}}$ is fixed, pruning remains dynamic because the model's continued learning causes new sequences to cross the threshold over time. The same upward shift in accuracy explains why the number of sequences below $\tau_{\mathrm{low}}$ decreases in the pruned distributions (Figure~\ref{fig:pruning_dynamics}(b,e)), even though pruning directly removes only sequences above $\tau_{\mathrm{low}}$.

\vspace{-5pt}
\paragraph{Pruning is safe at both extremes.}
If a sequence has temporarily low accuracy early in pretraining, when the model is still close to its initialization, it is simply retained because pruning acts only on high-accuracy sequences. As training progresses and its accuracy stabilizes, the sequence can still become eligible for pruning later. One pass over the corpus is largely sufficient for the sequence ranking to begin stabilizing. We verify this by measuring the rank correlation between sequence accuracy at a given epoch and its average accuracy over the final three epochs. Between the first and second epochs, this correlation rises from 0.36 to 0.63 for unpruned DB2 and from 0.75 to 0.93 for unpruned Hyena, and continues to increase thereafter (Appendix~\ref{app:ranking}). Thus, initially low accuracy does not permanently exclude a sequence from later pruning. At the other extreme, hard sequences remain far below $\tau_{\mathrm{low}}$ throughout pretraining and are therefore never at risk of removal under any parameter setting we test. Sequences near $\tau_{\mathrm{low}}$ are the main exception: a small epoch-to-epoch fluctuation (Figure~\ref{fig:pruning_dynamics}(a,b)), rather than genuine improvement, may move a sequence into the pruning range. Averaging accuracy over several epochs before applying the pruning criterion reduces this effect, but at the cost of some data savings (Table~\ref{tab:average x window}).

\vspace{-5pt}
\section{Related work}
\vspace{-5pt}
Data pruning methods differ in their scoring signal, computational cost, and refresh frequency. Static methods score examples once using difficulty or uncertainty
\citep{cho2025lightweightdatasetpruningtraining,xia2023moderate}, influence or generalization
\citep{tan2023datapruningmovingonesampleout,yang2023datasetpruningreducingtraining}, or
information-theoretic criteria
\citep{tan2025datapruninginformationmaximization,pmlr-v267-chen25q}. Because these scores are not updated, their rankings may become stale as training progresses. Dynamic methods refresh scores during training
\citep{kim2025adamoptimizationadaptivebatch,yang2025dynamicdataselectionmeets,
jin2026orderdp,azeemi2024languagemodeldrivendatapruning,
yu2025groupleveldataselectionefficient}, with \citet{zhou2026inconsistency} showing that stale decisions can bias the retained distribution. Influence-based methods estimate an example's effect through gradients, influence functions, or Bayesian approximations
\citep{yu2024matesmodelawaredataselection,nikdan2025efficientdataselectionscale,
gurbuz2025peaksselectingkeytraining,deng2023acceleratedmodeltrainingbayesian,
chhabra2024what}, but are expensive at scale and have been studied primarily for fine-tuning. Coreset and diversity methods instead optimize coverage
\citep{tong2025coreset,cheng2025efficient,zhang2024harnessingdiversityimportantdata,
nguyen2025minibatchcoresetsmemoryefficientlanguage,li2025massmathematicaldataselection,
du2025disentanglingrolesrepresentationselection},
while another line of work targets label noise, calibration, or spurious correlations
\citep{lin2023optimalsampleselectionuncertainty,vysogorets2025dropdistributionallyrobustdata,
park2023robustdatapruninglabel}. These latter approaches rely on labels or class structure, which are unavailable in self-supervised genomic pretraining.

At pretraining scale, data selection typically operates over documents, batches, or fixed subsets to satisfy distributional or compute objectives
\citep{xie2023dataselectionlanguagemodels,yin2025computeconstraineddataselection,
bai2025efficientpretrainingdataselection,zheng2025elfslabelfreecoresetselection,
shum2025predictivedataselectiondata,
he2026paserposttrainingdataselection}. These approaches often require a separate scoring procedure or reference model, and their selection decisions are made at a coarser granularity than individual sequences. To the best of our knowledge, none has been applied to DNA pretraining. SP instead dynamically prunes individual genomic sequences at every epoch boundary using a signal already produced by the model's forward pass, without requiring a separate reference model.

\vspace{-5pt}
\section{Conclusion and Limitations}
\label{sec:conclusion}
\vspace{-5pt}
We introduced data pruning to DNA pretraining. Per-sequence accuracy, obtained directly from the forward pass the model already performs, separates easy sequences from hard ones well enough to guide pruning. Removing easy sequences preserves most of the benefit of pretraining across five settings and both MLM and CLM objectives. Two control experiments further show that the gains depend on which sequences are removed, rather than on data reduction alone. The method is also simple to adopt: it requires no changes to the model architecture, pretraining objective, or optimizer.

Our method requires completing an epoch before pruning can begin, so extending it to single-pass pretraining would require additional changes. One possibility is to begin with a smaller corpus, allowing the same compute budget to cover multiple passes and giving the criterion an epoch boundary at which to act. Another is to make pruning decisions before a sequence is passed through the model. Our framework also treats pruning criteria as alternatives; combining them, such that a sequence is removed only when multiple criteria agree, is a natural direction for future work. Another direction is to adapt the learning-rate schedule to the reduced pretraining budget, allowing pruned runs to reach lower learning rates and potentially improve downstream performance at the same savings. Finally, the biological properties shared by pruned sequences remain an open question. Does the pruning criterion identify genuine redundancy in the genome, or does it instead reveal that DNA pretraining continues to spend computation on sequences the model has already learned?

\subsubsection*{AI Use Statement}
We used large language models for writing assistance only, including grammar
and phrasing. No part of the methodology, experiments, analysis, or code was
generated by AI tools.

\subsubsection*{Reproducibility Statement}
All corpora and benchmarks used here are publicly available, and our code is available at \url{https://github.com/FaiasPromit/pruning-dna-sequence}. Pretraining
configurations are in Appendix~\ref{app:pretraining_details}, downstream
protocols in Appendix~\ref{app:finetune}, and implementations of the compared
methods in Appendix~\ref{app:baselines}. All primary experiments use seed 42; Appendix G reports additional multi-seed variance experiments.

\subsubsection*{Ethics Statement}
This work uses publicly available reference genomes and established benchmarks.
It involves no human subjects and no private or identifiable data.

\subsubsection*{Acknowledgments}
Portions of this research were conducted with high performance computational resources provided by the Louisiana Optical Network Infrastructure (\url{http://www.loni.org}).

\bibliographystyle{iclr2027_conference}
\bibliography{iclr2027_conference}

\begin{thebibliography}{51}
\providecommand{\natexlab}[1]{#1}
\providecommand{\url}[1]{\texttt{#1}}
\expandafter\ifx\csname urlstyle\endcsname\relax
  \providecommand{\doi}[1]{doi: #1}\else
  \providecommand{\doi}{doi: \begingroup \urlstyle{rm}\Url}\fi

\bibitem[Alam et~al.(2026)Alam, Islam, and Islam]{alam2026batch}
Md~Mustakin Alam, Shaker Islam, and Aminul Islam.
\newblock Batch pruning by activation stability.
\newblock In \emph{The Fourteenth International Conference on Learning
  Representations}, 2026.
\newblock URL \url{https://openreview.net/forum?id=TUADW7db5n}.

\bibitem[Azeemi et~al.(2024)Azeemi, Qazi, and
  Raza]{azeemi2024languagemodeldrivendatapruning}
Abdul~Hameed Azeemi, Ihsan~Ayyub Qazi, and Agha~Ali Raza.
\newblock Language model-driven data pruning enables efficient active learning,
  2024.
\newblock URL \url{https://arxiv.org/abs/2410.04275}.

\bibitem[Bai et~al.(2025)Bai, Yang, Wong, Sun, Peng, Zhuang, Zhang, Wu, Qiu,
  Zhang, Yuan, and He]{bai2025efficientpretrainingdataselection}
Tianyi Bai, Ling Yang, Zhen~Hao Wong, Fupeng Sun, Jiahui Peng, Xinlin Zhuang,
  Chi Zhang, Lijun Wu, Jiantao Qiu, Wentao Zhang, Binhang Yuan, and Conghui He.
\newblock Efficient pretraining data selection for language models via
  multi-actor collaboration, 2025.
\newblock URL \url{https://arxiv.org/abs/2410.08102}.

\bibitem[Bengio et~al.(2009)Bengio, Louradour, Collobert, and
  Weston]{curriculum}
Yoshua Bengio, J\'{e}r\^{o}me Louradour, Ronan Collobert, and Jason Weston.
\newblock Curriculum learning.
\newblock ICML '09, pp.\  41–48, New York, NY, USA, 2009. Association for
  Computing Machinery.
\newblock ISBN 9781605585161.
\newblock \doi{10.1145/1553374.1553380}.
\newblock URL \url{https://doi.org/10.1145/1553374.1553380}.

\bibitem[Boshar et~al.(2025)Boshar, Evans, Tang, Picard, Adel, Lorbeer, Rajesh,
  Karch, Sidbon, Emms, Mendoza-Revilla, Al-Ani, Seitz, Schiff, Bornachot,
  Hernandez, Lopez, Laterre, Beguir, Koo, Kuleshov, Stark, de~Almeida, and
  Pierrot]{NTV3}
Sam Boshar, Benjamin Evans, Ziqi Tang, Armand Picard, Yanis Adel, Franziska~K.
  Lorbeer, Chandana Rajesh, Tristan Karch, Shawn Sidbon, David Emms, Javier
  Mendoza-Revilla, Fatimah Al-Ani, Evan Seitz, Yair Schiff, Yohan Bornachot,
  Ariana Hernandez, Marie Lopez, Alexandre Laterre, Karim Beguir, Peter Koo,
  Volodymyr Kuleshov, Alexander Stark, Bernardo~P. de~Almeida, and Thomas
  Pierrot.
\newblock A foundational model for joint sequence-function multi-species
  modeling at scale for long-range genomic prediction.
\newblock \emph{bioRxiv}, 2025.
\newblock \doi{10.64898/2025.12.22.695963}.
\newblock URL
  \url{https://www.biorxiv.org/content/early/2025/12/25/2025.12.22.695963}.

\bibitem[Brixi et~al.(2026)Brixi, Durrant, Ku, Naghipourfar, Poli, Sun,
  Brockman, Chang, Fanton, Gonzalez, King, Li, Merchant, Nguyen, Ricci-Tam,
  Romero, Schmok, Taghibakhshi, Vorontsov, Yang, et~al.]{evo2}
Garyk Brixi, Matthew~G. Durrant, Jerome Ku, Mohsen Naghipourfar, Michael Poli,
  Gwanggyu Sun, Greg Brockman, Daniel Chang, Alison Fanton, Gabriel~A.
  Gonzalez, Samuel~H. King, David~B. Li, Aditi~T. Merchant, Eric Nguyen, Chiara
  Ricci-Tam, David~W. Romero, Jonathan~C. Schmok, Ali Taghibakhshi, Anton
  Vorontsov, Brandon Yang, et~al.
\newblock Genome modelling and design across all domains of life with evo 2.
\newblock \emph{Nature}, 652\penalty0 (8112):\penalty0 1349--1361, 2026.
\newblock \doi{10.1038/s41586-026-10176-5}.

\bibitem[Chen(2025)]{pmlr-v267-chen25q}
Jianting Chen.
\newblock Efficient core-set selection for deep learning through squared loss
  minimization.
\newblock In Aarti Singh, Maryam Fazel, Daniel Hsu, Simon Lacoste-Julien, Felix
  Berkenkamp, Tegan Maharaj, Kiri Wagstaff, and Jerry Zhu (eds.),
  \emph{Proceedings of the 42nd International Conference on Machine Learning},
  volume 267 of \emph{Proceedings of Machine Learning Research}, pp.\
  8065--8076. PMLR, 13--19 Jul 2025.
\newblock URL \url{https://proceedings.mlr.press/v267/chen25q.html}.

\bibitem[Cheng et~al.(2025)Cheng, Wu, Li, Liao, Zhao, Chen, Gao, and
  ZHANG]{cheng2025efficient}
Zihao Cheng, Binrui Wu, Zhiwei Li, Yuesen Liao, Su~Zhao, Shuai Chen, Yuan Gao,
  and WEIZHONG ZHANG.
\newblock Efficient representativeness-aware coreset selection.
\newblock In \emph{The Thirty-ninth Annual Conference on Neural Information
  Processing Systems}, 2025.
\newblock URL \url{https://openreview.net/forum?id=CrdrVdXCui}.

\bibitem[Chhabra et~al.(2024)Chhabra, Li, Mohapatra, and Liu]{chhabra2024what}
Anshuman Chhabra, Peizhao Li, Prasant Mohapatra, and Hongfu Liu.
\newblock ''what data benefits my classifier?'' enhancing model performance and
  interpretability through influence-based data selection.
\newblock In \emph{The Twelfth International Conference on Learning
  Representations}, 2024.
\newblock URL \url{https://openreview.net/forum?id=HE9eUQlAvo}.

\bibitem[Cho et~al.(2025)Cho, Shin, Kang, and
  Yun]{cho2025lightweightdatasetpruningtraining}
Yeseul Cho, Baekrok Shin, Changmin Kang, and Chulhee Yun.
\newblock Lightweight dataset pruning without full training via example
  difficulty and prediction uncertainty, 2025.
\newblock URL \url{https://arxiv.org/abs/2502.06905}.

\bibitem[Dalla-Torre et~al.(2025)Dalla-Torre, Gonzalez, Mendoza-Revilla,
  et~al.]{dallatorre2025nucleotidetransformer}
Hugo Dalla-Torre, Liam Gonzalez, Javier Mendoza-Revilla, et~al.
\newblock Nucleotide transformer: Building and evaluating robust foundation
  models for human genomics.
\newblock \emph{Nature Methods}, 22:\penalty0 287--297, 2025.
\newblock \doi{10.1038/s41592-024-02523-z}.

\bibitem[Deng et~al.(2023)Deng, Cui, and
  Zhu]{deng2023acceleratedmodeltrainingbayesian}
Zhijie Deng, Peng Cui, and Jun Zhu.
\newblock Towards accelerated model training via bayesian data selection, 2023.
\newblock URL \url{https://arxiv.org/abs/2308.10544}.

\bibitem[Du et~al.(2025)Du, Song, Wong, Ignatev, Gatt, and
  Nguyen]{du2025disentanglingrolesrepresentationselection}
Yupei Du, Yingjin Song, Hugh~Mee Wong, Daniil Ignatev, Albert Gatt, and Dong
  Nguyen.
\newblock Disentangling the roles of representation and selection in data
  pruning, 2025.
\newblock URL \url{https://arxiv.org/abs/2507.03648}.

\bibitem[{ENCODE Project Consortium}(2020)]{encode2020expanded}
{ENCODE Project Consortium}.
\newblock Expanded encyclopaedias of {DNA} elements in the human and mouse
  genomes.
\newblock \emph{Nature}, 583\penalty0 (7818):\penalty0 699--710, 2020.
\newblock \doi{10.1038/s41586-020-2493-4}.

\bibitem[Fishman et~al.(2025)Fishman, Kuratov, Shmelev, Petrov, Penzar,
  Shepelin, Chekanov, Kardymon, and Burtsev]{10.1093/nar/gkae1310}
Veniamin Fishman, Yuri Kuratov, Aleksei Shmelev, Maxim Petrov, Dmitry Penzar,
  Denis Shepelin, Nikolay Chekanov, Olga Kardymon, and Mikhail Burtsev.
\newblock Gena-lm: a family of open-source foundational dna language models for
  long sequences.
\newblock \emph{Nucleic Acids Research}, 53\penalty0 (2):\penalty0 gkae1310, 01
  2025.
\newblock ISSN 1362-4962.
\newblock \doi{10.1093/nar/gkae1310}.
\newblock URL \url{https://doi.org/10.1093/nar/gkae1310}.

\bibitem[Gurbuz et~al.(2025)Gurbuz, Zheng, and
  Dovrolis]{gurbuz2025peaksselectingkeytraining}
Mustafa~Burak Gurbuz, Xingyu Zheng, and Constantine Dovrolis.
\newblock Peaks: Selecting key training examples incrementally via prediction
  error anchored by kernel similarity, 2025.
\newblock URL \url{https://arxiv.org/abs/2504.05250}.

\bibitem[He et~al.(2026)He, Yin, Zhen, Zhang, Yuan, and
  Ma]{he2026paserposttrainingdataselection}
Bowei He, Lihao Yin, Hui-Ling Zhen, Xiaokun Zhang, Mingxuan Yuan, and Chen Ma.
\newblock Paser: Post-training data selection for efficient pruned large
  language model recovery, 2026.
\newblock URL \url{https://arxiv.org/abs/2502.12594}.

\bibitem[Ji et~al.(2021)Ji, Zhou, Liu, and Davuluri]{ji2021dnabert}
Yanrong Ji, Zhihan Zhou, Han Liu, and Ramana~V Davuluri.
\newblock Dnabert: pre-trained bidirectional encoder representations from
  transformers model for dna-language in genome.
\newblock \emph{Bioinformatics}, 37\penalty0 (15):\penalty0 2112--2120, 08
  2021.
\newblock ISSN 1367-4803.
\newblock \doi{10.1093/bioinformatics/btab083}.
\newblock URL \url{https://doi.org/10.1093/bioinformatics/btab083}.

\bibitem[Jin et~al.(2026)Jin, Xu, Wang, JIA, Chen, and Zeng]{jin2026orderdp}
Chenhan Jin, Shengze Xu, Qingsong Wang, Fan JIA, Dingshuo Chen, and Tieyong
  Zeng.
\newblock Order{DP}: A theoretically guaranteed lossless dynamic data pruning
  framework.
\newblock In \emph{The Fourteenth International Conference on Learning
  Representations}, 2026.
\newblock URL \url{https://openreview.net/forum?id=e77QyyRQPz}.

\bibitem[Kim \& hwan Oh(2025)Kim and hwan
  Oh]{kim2025adamoptimizationadaptivebatch}
Gyu~Yeol Kim and Min hwan Oh.
\newblock Adam optimization with adaptive batch selection, 2025.
\newblock URL \url{https://arxiv.org/abs/2512.06795}.

\bibitem[Lander et~al.(2001)Lander, Linton, Birren, Nusbaum, Zody, Baldwin,
  Devon, Dewar, Doyle, FitzHugh, et~al.]{lander2001human}
Eric~S. Lander, Lauren~M. Linton, Bruce Birren, Chad Nusbaum, Michael~C. Zody,
  Jennifer Baldwin, Keri Devon, Ken Dewar, Michael Doyle, William FitzHugh,
  et~al.
\newblock Initial sequencing and analysis of the human genome.
\newblock \emph{Nature}, 409\penalty0 (6822):\penalty0 860--921, 2001.
\newblock \doi{10.1038/35057062}.

\bibitem[Li et~al.(2025)Li, Yu, Cui, Zhang, Zhou, Ye, and
  Zhang]{li2025massmathematicaldataselection}
Jiazheng Li, Lu~Yu, Qing Cui, Zhiqiang Zhang, Jun Zhou, Yanfang Ye, and Chuxu
  Zhang.
\newblock Mass: Mathematical data selection via skill graphs for pretraining
  large language models, 2025.
\newblock URL \url{https://arxiv.org/abs/2503.14917}.

\bibitem[Lin et~al.(2023)Lin, Liu, Ye, Lian, Yao, and
  Zhang]{lin2023optimalsampleselectionuncertainty}
Yong Lin, Chen Liu, Chenlu Ye, Qing Lian, Yuan Yao, and Tong Zhang.
\newblock Optimal sample selection through uncertainty estimation and its
  application in deep learning, 2023.
\newblock URL \url{https://arxiv.org/abs/2309.02476}.

\bibitem[Marin et~al.(2024)Marin, Teufel, Horlacher, Madsen, Pultz, Winther,
  and Boomsma]{BEND}
Frederikke~Isa Marin, Felix Teufel, Marc Horlacher, Dennis Madsen, Dennis
  Pultz, Ole Winther, and Wouter Boomsma.
\newblock {BEND}: Benchmarking {DNA} language models on biologically meaningful
  tasks.
\newblock In \emph{The Twelfth International Conference on Learning
  Representations}, 2024.
\newblock URL \url{https://openreview.net/forum?id=uKB4cFNQFg}.

\bibitem[Marion et~al.(2023)Marion, Üstün, Pozzobon, Wang, Fadaee, and
  Hooker]{marion2023moreinvestigatingdatapruning}
Max Marion, Ahmet Üstün, Luiza Pozzobon, Alex Wang, Marzieh Fadaee, and Sara
  Hooker.
\newblock When less is more: Investigating data pruning for pretraining llms at
  scale, 2023.
\newblock URL \url{https://arxiv.org/abs/2309.04564}.

\bibitem[Nguyen et~al.(2025)Nguyen, Yang, Anand, Yang, and
  Mirzasoleiman]{nguyen2025minibatchcoresetsmemoryefficientlanguage}
Dang Nguyen, Wenhan Yang, Rathul Anand, Yu~Yang, and Baharan Mirzasoleiman.
\newblock Mini-batch coresets for memory-efficient language model training on
  data mixtures, 2025.
\newblock URL \url{https://arxiv.org/abs/2407.19580}.

\bibitem[Nguyen et~al.(2023)Nguyen, Poli, Faizi, Thomas, Wornow, Birch-Sykes,
  Massaroli, Patel, Rabideau, Bengio, Ermon, Re, and Baccus]{hyenadna}
Eric Nguyen, Michael Poli, Marjan Faizi, Armin~W Thomas, Michael Wornow, Callum
  Birch-Sykes, Stefano Massaroli, Aman Patel, Clayton~M. Rabideau, Yoshua
  Bengio, Stefano Ermon, Christopher Re, and Stephen Baccus.
\newblock Hyena{DNA}: Long-range genomic sequence modeling at single nucleotide
  resolution.
\newblock In \emph{Thirty-seventh Conference on Neural Information Processing
  Systems}, 2023.
\newblock URL \url{https://openreview.net/forum?id=ubzNoJjOKj}.

\bibitem[Nguyen et~al.(2024)Nguyen, Poli, Durrant, Kang, Katrekar, Li, Bartie,
  Thomas, King, Brixi, Sullivan, Ng, Lewis, Lou, Ermon, Baccus,
  Hernandez-Boussard, Ré, Hsu, and Hie]{evo}
Eric Nguyen, Michael Poli, Matthew~G. Durrant, Brian Kang, Dhruva Katrekar,
  David~B. Li, Liam~J. Bartie, Armin~W. Thomas, Samuel~H. King, Garyk Brixi,
  Jeremy Sullivan, Madelena~Y. Ng, Ashley Lewis, Aaron Lou, Stefano Ermon,
  Stephen~A. Baccus, Tina Hernandez-Boussard, Christopher Ré, Patrick~D. Hsu,
  and Brian~L. Hie.
\newblock Sequence modeling and design from molecular to genome scale with evo.
\newblock \emph{Science}, 386\penalty0 (6723):\penalty0 eado9336, 2024.
\newblock \doi{10.1126/science.ado9336}.
\newblock URL \url{https://www.science.org/doi/abs/10.1126/science.ado9336}.

\bibitem[Nikdan et~al.(2025)Nikdan, Cohen-Addad, Alistarh, and
  Mirrokni]{nikdan2025efficientdataselectionscale}
Mahdi Nikdan, Vincent Cohen-Addad, Dan Alistarh, and Vahab Mirrokni.
\newblock Efficient data selection at scale via influence distillation, 2025.
\newblock URL \url{https://arxiv.org/abs/2505.19051}.

\bibitem[Park et~al.(2023)Park, Choi, Kim, Song, and
  Lee]{park2023robustdatapruninglabel}
Dongmin Park, Seola Choi, Doyoung Kim, Hwanjun Song, and Jae-Gil Lee.
\newblock Robust data pruning under label noise via maximizing re-labeling
  accuracy, 2023.
\newblock URL \url{https://arxiv.org/abs/2311.01002}.

\bibitem[Patel et~al.(2024)Patel, Singhal, Wang, Pampari, Kasowski, and
  Kundaje]{DART-eval}
Aman Patel, Arpita Singhal, Austin Wang, Anusri Pampari, Maya Kasowski, and
  Anshul Kundaje.
\newblock {DART}-eval: A comprehensive {DNA} language model evaluation
  benchmark on regulatory {DNA}.
\newblock In \emph{The Thirty-eight Conference on Neural Information Processing
  Systems Datasets and Benchmarks Track}, 2024.
\newblock URL \url{https://openreview.net/forum?id=qR0x6H5WUX}.

\bibitem[Paul et~al.(2021)Paul, Ganguli, and Dziugaite]{el2n}
Mansheej Paul, Surya Ganguli, and Gintare~Karolina Dziugaite.
\newblock Deep learning on a data diet: Finding important examples early in
  training.
\newblock In A.~Beygelzimer, Y.~Dauphin, P.~Liang, and J.~Wortman Vaughan
  (eds.), \emph{Advances in Neural Information Processing Systems}, 2021.
\newblock URL \url{https://openreview.net/forum?id=Uj7pF-D-YvT}.

\bibitem[Qin et~al.(2024)Qin, Wang, Zheng, Gu, Peng, xu~Zhao~Pan, Zhou, Shang,
  Sun, Xie, and You]{infobatch}
Ziheng Qin, Kai Wang, Zangwei Zheng, Jianyang Gu, Xiangyu Peng, xu~Zhao~Pan,
  Daquan Zhou, Lei Shang, Baigui Sun, Xuansong Xie, and Yang You.
\newblock Infobatch: Lossless training speed up by unbiased dynamic data
  pruning.
\newblock In \emph{The Twelfth International Conference on Learning
  Representations}, 2024.
\newblock URL \url{https://openreview.net/forum?id=C61sk5LsK6}.

\bibitem[Shum et~al.(2025)Shum, Huang, Zou, Ding, Liao, Chen, Liu, and
  He]{shum2025predictivedataselectiondata}
Kashun Shum, Yuzhen Huang, Hongjian Zou, Qi~Ding, Yixuan Liao, Xiaoxin Chen,
  Qian Liu, and Junxian He.
\newblock Predictive data selection: The data that predicts is the data that
  teaches, 2025.
\newblock URL \url{https://arxiv.org/abs/2503.00808}.

\bibitem[Tan et~al.(2023)Tan, Wu, Du, Chen, Wang, Wang, and
  Qi]{tan2023datapruningmovingonesampleout}
Haoru Tan, Sitong Wu, Fei Du, Yukang Chen, Zhibin Wang, Fan Wang, and Xiaojuan
  Qi.
\newblock Data pruning via moving-one-sample-out, 2023.
\newblock URL \url{https://arxiv.org/abs/2310.14664}.

\bibitem[Tan et~al.(2025)Tan, Wu, Huang, Zhao, and
  Qi]{tan2025datapruninginformationmaximization}
Haoru Tan, Sitong Wu, Wei Huang, Shizhen Zhao, and Xiaojuan Qi.
\newblock Data pruning by information maximization, 2025.
\newblock URL \url{https://arxiv.org/abs/2506.01701}.

\bibitem[Tong et~al.(2025)Tong, Liu, Shi, and Gong]{tong2025coreset}
Ruilin Tong, Yuhang Liu, Javen~Qinfeng Shi, and Dong Gong.
\newblock Coreset selection via reducible loss in continual learning.
\newblock In \emph{The Thirteenth International Conference on Learning
  Representations}, 2025.
\newblock URL \url{https://openreview.net/forum?id=mAztx8QO3B}.

\bibitem[Vishniakov et~al.(2026)Vishniakov, Viswanathan, Medvedev, Kanithi,
  Pimentel, Rajan, and Khan]{nopretrain}
Kirill Vishniakov, Karthik Viswanathan, Aleksandr Medvedev, Praveenkumar
  Kanithi, Marco~AF Pimentel, Ronnie Rajan, and Shadab Khan.
\newblock Tokenization to transfer: Do genomic foundation models learn good
  representations?
\newblock In \emph{The Fourteenth International Conference on Learning
  Representations}, 2026.
\newblock URL \url{https://openreview.net/forum?id=4UY1NHG5Ge}.

\bibitem[Vysogorets et~al.(2025)Vysogorets, Ahuja, and
  Kempe]{vysogorets2025dropdistributionallyrobustdata}
Artem Vysogorets, Kartik Ahuja, and Julia Kempe.
\newblock Drop: Distributionally robust data pruning, 2025.
\newblock URL \url{https://arxiv.org/abs/2404.05579}.

\bibitem[Wu et~al.(2026)Wu, Jiang, Ye, Wang, Zhou, Xu, Chen, Hu, Zhang, Jin,
  Yuan, and Li]{wu2026investigatingdatapruningpretraining}
Yifan Wu, Jiyue Jiang, Xichen Ye, Yiqi Wang, Chang Zhou, Yitao Xu, Jiayang
  Chen, He~Hu, Weizhong Zhang, Cheng Jin, Jiao Yuan, and Yu~Li.
\newblock Investigating data pruning for pretraining biological foundation
  models at scale, 2026.
\newblock URL \url{https://arxiv.org/abs/2512.12932}.

\bibitem[Xia et~al.(2023)Xia, Liu, Yu, Shen, Han, and Liu]{xia2023moderate}
Xiaobo Xia, Jiale Liu, Jun Yu, Xu~Shen, Bo~Han, and Tongliang Liu.
\newblock Moderate coreset: A universal method of data selection for real-world
  data-efficient deep learning.
\newblock In \emph{The Eleventh International Conference on Learning
  Representations}, 2023.
\newblock URL \url{https://openreview.net/forum?id=7D5EECbOaf9}.

\bibitem[Xie et~al.(2023)Xie, Santurkar, Ma, and
  Liang]{xie2023dataselectionlanguagemodels}
Sang~Michael Xie, Shibani Santurkar, Tengyu Ma, and Percy Liang.
\newblock Data selection for language models via importance resampling, 2023.
\newblock URL \url{https://arxiv.org/abs/2302.03169}.

\bibitem[Yang et~al.(2023)Yang, Xie, Peng, Xu, Sun, and
  Li]{yang2023datasetpruningreducingtraining}
Shuo Yang, Zeke Xie, Hanyu Peng, Min Xu, Mingming Sun, and Ping Li.
\newblock Dataset pruning: Reducing training data by examining generalization
  influence, 2023.
\newblock URL \url{https://arxiv.org/abs/2205.09329}.

\bibitem[Yang et~al.(2025)Yang, Ye, Shen, and
  Zhou]{yang2025dynamicdataselectionmeets}
Suorong Yang, Peng Ye, Furao Shen, and Dongzhan Zhou.
\newblock When dynamic data selection meets data augmentation, 2025.
\newblock URL \url{https://arxiv.org/abs/2505.03809}.

\bibitem[Yin \& Rush(2025)Yin and Rush]{yin2025computeconstraineddataselection}
Junjie~Oscar Yin and Alexander~M. Rush.
\newblock Compute-constrained data selection, 2025.
\newblock URL \url{https://arxiv.org/abs/2410.16208}.

\bibitem[Yu et~al.(2024)Yu, Das, and Xiong]{yu2024matesmodelawaredataselection}
Zichun Yu, Spandan Das, and Chenyan Xiong.
\newblock Mates: Model-aware data selection for efficient pretraining with data
  influence models, 2024.
\newblock URL \url{https://arxiv.org/abs/2406.06046}.

\bibitem[Yu et~al.(2025)Yu, Peng, Lei, Overwijk, tau Yih, and
  Xiong]{yu2025groupleveldataselectionefficient}
Zichun Yu, Fei Peng, Jie Lei, Arnold Overwijk, Wen tau Yih, and Chenyan Xiong.
\newblock Group-level data selection for efficient pretraining, 2025.
\newblock URL \url{https://arxiv.org/abs/2502.14709}.

\bibitem[Zhang et~al.(2024)Zhang, Zhong, Zhang, Chai, Wang, Zhuang, Bai, Qiu,
  Cao, Fan, Yuan, Wang, and He]{zhang2024harnessingdiversityimportantdata}
Chi Zhang, Huaping Zhong, Kuan Zhang, Chengliang Chai, Rui Wang, Xinlin Zhuang,
  Tianyi Bai, Jiantao Qiu, Lei Cao, Ju~Fan, Ye~Yuan, Guoren Wang, and Conghui
  He.
\newblock Harnessing diversity for important data selection in pretraining
  large language models, 2024.
\newblock URL \url{https://arxiv.org/abs/2409.16986}.

\bibitem[Zheng et~al.(2025)Zheng, Tsai, Lu, Sun, Bartoldson, Kailkhura, and
  Prakash]{zheng2025elfslabelfreecoresetselection}
Haizhong Zheng, Elisa Tsai, Yifu Lu, Jiachen Sun, Brian~R. Bartoldson, Bhavya
  Kailkhura, and Atul Prakash.
\newblock Elfs: Label-free coreset selection with proxy training dynamics,
  2025.
\newblock URL \url{https://arxiv.org/abs/2406.04273}.

\bibitem[Zhou et~al.(2026)Zhou, Yang, Zhao, Zhang, Gao, and
  Wang]{zhou2026inconsistency}
Qing Zhou, Tao Yang, Bingxuan Zhao, Hongyuan Zhang, Junyu Gao, and Qi~Wang.
\newblock Inconsistency biases in dynamic data pruning.
\newblock In \emph{The Fourteenth International Conference on Learning
  Representations}, 2026.
\newblock URL \url{https://openreview.net/forum?id=Zw1Uw7u6Su}.

\bibitem[Zhou et~al.(2024)Zhou, Ji, Li, Dutta, Davuluri, and
  Liu]{zhou2024dnabert2}
Zhihan Zhou, Yanrong Ji, Weijian Li, Pratik Dutta, Ramana~V Davuluri, and Han
  Liu.
\newblock {DNABERT}-2: Efficient foundation model and benchmark for
  multi-species genomes.
\newblock In \emph{The Twelfth International Conference on Learning
  Representations}, 2024.
\newblock URL \url{https://openreview.net/forum?id=oMLQB4EZE1}.

\end{thebibliography}

\appendix
\FloatBarrier

\par
\Needspace{10\baselineskip}
\begin{wraptable}{R}{0.59\textwidth}
\renewcommand{\tabcolsep}{1pt}
\centering \vspace{-5pt}
\vspace{-5pt}\caption{Per-sequence accuracy across the unpruned baselines, over every sequence in each corpus. Fluctuation is the standard deviation of a sequence's accuracy across epochs, reported as the median over sequences. Accuracy is bounded by 0 and 1 in both settings.}\vspace{-5pt}
\label{tab:stability}
\begin{tabular}{lcc}
\toprule
 & DB2--MSG & Hyena--HRG \\
 & (MLM) & (CLM) \\
\midrule
Early window, mean accuracy  & 0.233 & 0.530 \\
Late window, mean accuracy   & 0.274 & 0.551 \\
Change in mean               & $+0.042$ & $+0.020$ \\
Sequences that improved      & 74.8\% & 84.2\% \\
\midrule
Fluctuation across epochs    & 0.096 & 0.009 \\
\bottomrule
\end{tabular}
\end{wraptable}

\section{Per-Sequence Accuracy as a Difficulty Signal}
\label{app:accuracy}

This appendix provides additional evidence supporting the claims associated with Figure~\ref{fig:acc_samples} and~\ref{fig:acc_corpus}. For both unpruned baselines, we track every sequence in the corpus at every epoch: 32.4M MSG sequences for DB2 over 47 epochs and 2.7M HRG sequences for Hyena over 28 epochs. When comparing early and late training windows, we use the first and last twenty epochs for DB2 and the first and last ten epochs for HyenaDNA, reflecting the different run lengths.

\subsection{Sequences Stay Within a Band}
\label{app:stability}

A sequence's accuracy changes slightly from one epoch to the next, as also shown in Figure~\ref{fig:acc_samples}. On a 0--1 accuracy scale, the median fluctuation in per-sequence accuracy is 0.096 for DB2 and 0.009 for Hyena (Table~\ref{tab:stability}). These fluctuations are small enough that a sequence typically remains within the same accuracy band throughout pretraining.

\par
\Needspace{20\baselineskip}
\begin{wraptable}{R}{0.40\textwidth}
\renewcommand{\tabcolsep}{1pt}
\centering \vspace{-5pt}
\vspace{-5pt} \caption{Spearman rank correlation between per-sequence accuracy at a single epoch and each sequence's eventual difficulty, taken as its mean accuracy over the final three epochs.} \vspace{-5pt}
\label{tab:ranking}
\begin{tabular}{lcc}
\toprule
\multirow{2}{*}{Epoch} & DB2 (MLM) & Hyena (CLM) \\
 & $\rho$ & $\rho$ \\
\midrule
1  & 0.36 & 0.75 \\
2  & 0.63 & 0.93 \\
3  & 0.68 & 0.96 \\
5  & 0.72 & 0.98 \\
10 & 0.77 & 0.99 \\
20 & 0.81 & 1.00 \\
\bottomrule
\end{tabular} \vspace{-15pt}
\end{wraptable}

\subsection{Continued Training Improves Accuracy Slightly}

From the early to the late window, mean accuracy increases by 4.2 percentage points for DB2 and 2.0 percentage points for Hyena (Table~\ref{tab:stability}). Accuracy improves for 74.8\% and 84.2\% of sequences, respectively, showing that both models continue to learn throughout pretraining. However, the magnitude of this improvement is small, so sequences generally remain within similar accuracy bands. As a result, accuracy continues to separate easy sequences from hard ones even as the model improves.

\subsection{Accuracy Recovers the Eventual Ordering Early}
\label{app:ranking}

Beyond separating easy sequences from hard ones, per-sequence pretraining accuracy also ranks sequences by difficulty, and this ordering becomes reliable early enough to guide pruning. We use each sequence's mean accuracy over the final three epochs as its reference difficulty. We then compare this ranking with the ranking obtained from a single earlier epoch using Spearman's rank correlation $\rho$. A value of 1 indicates identical rankings, while 0 indicates no relation between them. Because this measure depends only on the relative ordering of sequences, the overall increase in accuracy during pretraining does not affect it.

Both settings exhibit the same pattern (Table~\ref{tab:ranking}). At epoch~1, accuracy provides only a weak estimate of the eventual ordering, consistent with the model still being close to its initialization. After one additional pass over the corpus, agreement increases substantially and continues to strengthen as pretraining proceeds. By the time most pruning occurs, the ordering from a single epoch already closely agrees with the eventual ordering, supporting its use as the pruning criterion.

\par
\Needspace{10\baselineskip}
\begin{wrapfigure}{R}{0.65\textwidth}
    \centering
    \includegraphics[width=\linewidth]{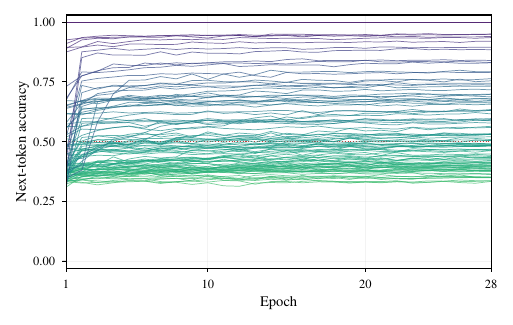}
    \caption{100 randomly sampled sequences from the Hyena--HRG unpruned baseline. Colour encodes each sequence's mean accuracy. The dotted line marks 0.5 for reference.}
    \label{fig:all_hyena}
\end{wrapfigure}

\subsection{The Two Objectives Differ in Measurement Precision}
\label{app:clm_mlm}

The stability measures in Appendix~\ref{app:accuracy} are consistently stronger for HyenaDNA than for DB2: the median fluctuation is 0.009 compared with 0.096, and the rank correlations are consistently higher. The same pattern appears in Figure~\ref{fig:acc_samples}, where the trajectories in panel~(b) are nearly flat, while those in panel~(a) fluctuate around their typical levels. CLM accuracy is computed over every prediction position in a sequence, whereas MLM accuracy uses only the masked positions, which may comprise only a few positions in a short sequence. Moreover, the same sequence receives a different mask at each epoch, so its MLM accuracy may be computed from different positions over time. CLM therefore uses more prediction positions per sequence and yields a more stable accuracy estimate. The two objectives differ in how precisely per-sequence accuracy is measured, not in whether accuracy separates easy and hard sequences.

Because Hyena trajectories are highly stable, 100 randomly sampled sequences can be displayed together while remaining easy to distinguish. Figure~\ref{fig:all_hyena} shows that each sequence remains at a similar accuracy level throughout pretraining, with most crossings occurring during the first few epochs, when the model is still close to its initialization.

\FloatBarrier

\section{Additional Method Details}
\label{app:framework_details}

This appendix provides additional methodological details for the pruning framework illustrated in Figure~\ref{fig:pipeline}. We use the same notation as in Section~2: $N$ is the number of sequences, $p$ indexes a sequence, $I$ is the scheduled number of epochs, $i$ indexes an epoch, and $\mathcal{A}_i \subseteq \{1,\ldots,N\}$ is the active corpus entering epoch $i$. The batch size is $b$, and $\ell$ denotes the number of model blocks. For sequence $p$, $M_{p,\mathrm{total}}^{(i)}$ denotes the number of scored positions in epoch $i$, and $\mathrm{Acc}_p^{(i)}$ denotes its per-sequence accuracy when $M_{p,\mathrm{total}}^{(i)}>0$. We first provide additional details of the SP framework, followed by InfoBatch and BPAS, and then the perplexity and EL2N variants.

\subsection{SP Framework Details}

The main method defines the pruning criterion; this subsection provides the additional details needed to maintain $\mathcal{A}_i$ across epochs and handle sequences for which no valid score is available. Sequence identifiers are preserved across epochs so that each sequence can be tracked as $\mathcal{A}_i$ changes and batches are reconstructed. This ensures that sequences are correctly retained or excluded after each pruning decision. Once sequence $p$ is removed, it does not return to $\mathcal{A}_{i+1}$ or any later active corpus.

At each epoch, the sequences in $\mathcal{A}_i$ are shuffled and regrouped into batches of size $b$. If $|\mathcal{A}_i|$ is not divisible by $b$, the remaining sequences do not form a complete batch and are therefore not processed or scored in epoch $i$. Because $\mathcal{A}_i$ is reshuffled and rebatched at the next epoch, the sequences in this remainder may differ across epochs. Thus, a sequence omitted in one epoch can be processed in a later epoch, preventing the same subset of the corpus from being systematically excluded from the pruning criterion. Under MLM, a processed sequence may also receive no \texttt{[MASK]} positions, in which case $M_{p,\mathrm{total}}^{(i)}=0$ and $\mathrm{Acc}_p^{(i)}$ is undefined. Such sequences are excluded from the pruning decision for epoch $i$ and reconsidered in a later epoch.

\subsection{Comparison Methods and Alternative Pruning Criteria}
\label{app:baselines}

We run InfoBatch and BPAS using the same corpus, step budget, and termination rule as SP in the corresponding pretraining setting. Perplexity and EL2N instead replace the pruning criterion used in block B of Figure~\ref{fig:pipeline}, while the remaining SP procedure and step budget remain unchanged.

\subsubsection{InfoBatch}
\label{app:infobatch}

Following \citet{infobatch}, InfoBatch dynamically scores sequence $p$ by its mean cross-entropy loss over epoch $i$ and temporarily drops low-loss sequences rather than removing them permanently. At each epoch, the mean of all $N$ sequence scores is used as the threshold. Sequences at or above the threshold are always retained, while those below it are independently dropped with probability $r$. Selection is resampled at every epoch, so no sequence is permanently removed.

To correct the gradient bias introduced by dropping part of the low-loss group, the loss of retained sequences from that group is scaled by $1/(1-r)$, restoring the expected full-data gradient. Pruning is applied only during the first $\delta I$ epochs; afterward, all $N$ sequences are pretrained without rescaling. Scores are initialized to $1$ before epoch $1$, and the score of sequence $p$ is updated only in epochs in which $p$ is processed; otherwise, its previous score is retained. Table~\ref{tab:variant-parameters} gives $(r,\delta)$ for the InfoBatch variants.

\subsubsection{BPAS}
\label{app:bpas}

Following \citet{alam2026batch}, BPAS partitions the $N$ sequences into fixed global batches before pretraining begins. Each sequence remains assigned to the same batch throughout the run. Sequences are shuffled within each batch, and the order of batches is shuffled, but sequences never move between batches. The trailing incomplete batch is dropped so that all global batches have the same size.

For each batch and transformer block, we flatten the block output and record its standard deviation. Averaging these values over the $\ell$ blocks produces one statistic per batch at epoch $i$. From epoch $2$ onward, each active batch is compared with its statistic from epoch $i-1$ and pruned when the change falls below $\delta(i) = \delta_s e^{\alpha i}$, where $\alpha = \tfrac{1}{I}\ln(\delta_e/\delta_s)$. The threshold is more restrictive early in pretraining, when representations are changing rapidly, and relaxes as pretraining progresses. Once a batch is pruned, all of its sequences are permanently removed, causing $\mathcal{A}_i$ to shrink monotonically. Table~\ref{tab:variant-parameters} gives $(\delta_s,\delta_e)$ for each BPAS variant.

\subsubsection{Perplexity}
\label{app:perp}

Following \citet{marion2023moreinvestigatingdatapruning}, perplexity scores sequence $p$ at epoch $i$ as the exponential of the mean negative log-likelihood over its prediction positions. Perplexity is bounded below by $1$, attained when every position is predicted with certainty, and is unbounded above.

Low perplexity therefore identifies sequences that are already predicted well. The low threshold is fixed at $1$, while the high threshold varies across variants; sequence $p$ is pruned when its perplexity is at or below the high threshold. The score is obtained from the same forward pass, pruning is applied at the epoch boundary, and removed sequences remain excluded from subsequent $\mathcal{A}_i$. A sequence without a valid score in epoch $i$ is excluded from that epoch's pruning decision and reconsidered in a later epoch. Table~\ref{tab:variant-parameters} gives the high threshold for each perplexity variant.

\subsubsection{EL2N}
\label{app:el2n}

Following \citet{el2n}, EL2N scores sequence $p$ at epoch $i$ by the  \(\ell_2\) norm of its prediction error. At each masked position, we compute the norm between the predicted softmax distribution and the one-hot target and average these values over the masked positions of $p$. The score lies in $[0,\sqrt{2}]$, where $0$ corresponds to a perfect prediction, a value near $1$ to a uniform prediction over the vocabulary, and $\sqrt{2}$ to a confident incorrect prediction.

Low EL2N therefore identifies sequences that are already predicted well. Its threshold orientation is opposite to that of accuracy: accuracy fixes $\tau_{\mathrm{high}}=1$ and varies $\tau_{\mathrm{low}}$, whereas EL2N fixes its low threshold at $0$ and varies its high threshold. The score is obtained from the same forward pass, pruning is applied at the epoch boundary, and removed sequences remain excluded from subsequent $\mathcal{A}_i$. If $M_{p,\mathrm{total}}^{(i)}=0$, sequence $p$ is excluded from the pruning decision for epoch $i$ and reconsidered in a later epoch. Table~\ref{tab:variant-parameters} gives the high threshold for each EL2N variant.

\par
\Needspace{20\baselineskip}
\begin{wraptable}{R}{0.70\textwidth}
\renewcommand{\tabcolsep}{1pt}
\centering\vspace{-5pt}
\small
\vspace{-5pt}\caption{Optimizer hyperparameters and learning-rate schedules for each pretraining setting. Hyena denotes the 1,024-bp Hyena setting.}\vspace{-5pt}
\label{tab:hparams}
\begin{adjustbox}{max width=\linewidth}
\begin{tabular}{lccccc}
\toprule
 & DB2 & DB2 & NT-500M & Hyena & Hyena-16k \\
 & MSG & HRG & HRG & HRG & HRG \\
\midrule
Optimizer       & AdamW & AdamW & Adam & AdamW & AdamW \\
Peak LR         & $5\!\times\!10^{-4}$ & $5\!\times\!10^{-4}$ & $1\!\times\!10^{-4}$ & $6\!\times\!10^{-4}$ & $6\!\times\!10^{-4}$ \\
LR Schedule     & linear & linear & inverse sqrt & cosine & cosine \\
Warmup steps    & 30{,}000 & 1{,}800 & 16{,}000 & 500 & 500 \\
Adam $\beta$    & (0.9, 0.98) & (0.9, 0.98) & (0.9, 0.999) & (0.9, 0.999) & (0.9, 0.999) \\
Adam $\epsilon$ & $10^{-6}$ & $10^{-6}$ & $10^{-8}$ & $10^{-8}$ & $10^{-8}$ \\
Weight decay    & $10^{-5}$ & $10^{-5}$ & 0 & 0.1 & 0.1 \\
\bottomrule
\end{tabular} 
\end{adjustbox} \vspace{-25pt}
\end{wraptable}

\FloatBarrier
\section{Pretraining Details}
\label{app:pretraining_details}

Table~\ref{tab:hparams} gives the optimizer and learning-rate schedule for
every setting. All runs train in bf16 with gradient clipping at $1.0$. The
subsections below give the corpus, tokenizer, and model configuration for each
setting in turn.

\subsection{DB2 on MSG}
\label{app:db2-msg}
The multi-species genome collection, segmented into
sequences of roughly $1$~kb. Tokenization uses byte-pair encoding with a
$4{,}096$-token vocabulary, under which a $1$~kb sequence becomes roughly 128
tokens. Ambiguous bases are dropped during corpus preparation. The model is the
released 117M-parameter DB2 architecture, pretrained from scratch with $15\%$ MLM
masking under the standard $80/10/10$ corruption rule.

\subsection{DB2 on HRG}
\label{app:db2-hrg}
The human reference genome, segmented into $1$~kb sequences and prepared with
the same pipeline as MSG: the same byte-pair tokenizer and vocabulary size, and
ambiguous bases dropped. The model and masking are identical to
Appendix~\ref{app:db2-msg}. This setting differs from that one only in corpus,
which is what isolates the effect of the data.

\subsection{NT-500M on HRG}
\label{app:nt-hrg}
The HRG, segmented into $6{,}100$~bp windows with $50$~bp of
overlap on each side, following the NT-500M protocol. Tokenization uses
non-overlapping 6-mers with a $4{,}105$-token vocabulary. Ambiguous bases are retained, since the
vocabulary carries an explicit \texttt{N}. The model follows the
NT-500M-HumanRef configuration, with 24 layers, hidden size 1280, 20 attention
heads, and intermediate size 5120, pretrained from scratch with $15\%$ masking
under the same $80/10/10$ rule. Its learning rate warms up linearly from
$5\!\times\!10^{-5}$ to peak, then decays as the inverse square root of the
step count with a floor at one percent of peak.

\subsection{Hyena on HRG}
\label{app:hyena}
The human reference genome, segmented into fixed windows of $1{,}024$~bp for
Hyena and $16{,}384$~bp for Hyena-16k. Tokenization is single-character with
a 12-token vocabulary, so tokens and base pairs correspond one to one, and
ambiguous bases are retained as an explicit \texttt{N} token. Both settings use the 8-layer, width-256 architecture with Hyena order 2 and implicit filter order 64, pretrained from scratch on next-token prediction. They share an optimizer and differ only in context length and step budget. Parameter counts differ slightly, 6.6M at 
1,024~bp and 7.2M at 16,384~bp, because each Hyena filter carries a learned positional embedding sized to the context length. Every other weight is identical between the two. We hold the context fixed at the stated length rather than applying the
sequence-length warmup of the original recipe. This keeps a sequence the same
length at every epoch, so its per-sequence accuracy stays comparable across
epochs.

\par
\Needspace{10\baselineskip}
\begin{wraptable}{R}{0.65\textwidth}
\renewcommand{\tabcolsep}{1pt}
\centering
\small
\vspace{-5pt}\caption{Fine-tuning settings by benchmark.}\vspace{-5pt}
\label{tab:ft}
\begin{adjustbox}{max width=\linewidth}
\begin{tabular}{lccc}
\toprule
& GUE & NT-18 & GB \\
\midrule
Pretrained models & DB2, NT-500M & DB2, NT-500M & Hyena \\
Tasks             & 27 tasks, 6 groups & 18 tasks, 3 groups & 7 tasks \\
Learning rate     & $3\!\times\!10^{-5}$ & $3\!\times\!10^{-5}$ & $6\!\times\!10^{-4}$ \\
Batch size        & 16 & 16 & 128 \\
Weight decay      & 0.01 & 0.01 & 0.1 \\
Warmup            & 100 steps & 100 steps & 1\% of steps \\
Epochs            & per task & per task & 20 \\
\bottomrule
\end{tabular}
\end{adjustbox}
\end{wraptable}

\FloatBarrier
\section{Downstream Protocols}
\label{app:finetune}

Our downstream evaluation uses two protocols: full-parameter fine-tuning for GUE, NT-18, and GB, and frozen probing for B-GF and D-T4. This appendix provides the training and evaluation details for both settings.

\subsection{Fine-tuned benchmarks}

We follow each benchmark's published fine-tuning protocol and we omit the optional Low-Rank Adaptation component and update all parameters. Table~\ref{tab:ft} gives the settings shared across tasks within a benchmark,
and Table~\ref{tab:ft-tasks} the per-task epochs and maximum sequence lengths.
Lengths follow each task's own sequence-length distribution, so they differ
between tokenizers at equal base-pair length. Every run uses seed 42, and we evaluate the best checkpoint by validation
performance: lowest validation loss on GUE and GB, and highest validation MCC
on NT-18.

\begin{table}[h]
\centering
\small
\setlength{\tabcolsep}{5pt}
\caption{Per-task training epochs and maximum token lengths for the fine-tuned benchmarks. TF denotes transcription factor.}
\label{tab:ft-tasks}
\begin{adjustbox}{max width=\textwidth}
\begin{tabular}{llccc}
\toprule
\multirow{2}{*}{Task group}
 & \multirow{2}{*}{Task}
 & \multirow{2}{*}{Epochs}
 & \multicolumn{2}{c}{Maximum token length} \\
\cmidrule(lr){4-5}
 & & & DB2 & NT-500M \\
\midrule
\multicolumn{5}{l}{\textbf{GUE}} \\
Epigenetic marks  & 10 histone datasets & 3  & 132 & 112 \\
TF binding, mouse & 5 datasets          & 30 & 47  & 24  \\
TF binding, human & 5 datasets          & 3  & 29  & 24  \\
\cmidrule(lr){2-5}
\multirow{3}{*}{Promoter} & all            & 4  & 77  & 64  \\
                  & no TATA             & 4  & 77  & 64  \\
                  & TATA                & 10 & 82  & 64  \\
\cmidrule(lr){2-5}
\multirow{3}{*}{Core Promoter} & all       & 4  & 21  & 16  \\
                  & no TATA             & 4  & 21  & 16  \\
                  & TATA                & 10 & 27  & 20  \\
\cmidrule(lr){2-5}
Splice site       & reconstruction      & 5  & 110 & 96  \\
\midrule
\multicolumn{5}{l}{\textbf{NT-18}} \\
\multirow{5}{*}{Regulatory} & promoter (all)           & 4  & 100 & 64  \\
            & promoter (no TATA)                   & 4  & 100 & 64  \\
            & promoter (TATA)                      & 10 & 100 & 64  \\
            & enhancers     & 4  & 128 & 96  \\
            & enhancer types            & 4  & 128 & 96  \\
\cmidrule(lr){2-5}
\multirow{3}{*}{Splice site} & all        & 5  & 192 & 128 \\
            & acceptors                 & 5  & 192 & 128 \\
            & donors                    & 5  & 192 & 128 \\
\cmidrule(lr){2-5}
Chromatin   & 10 histone datasets        & 3  & 256 & 192 \\
\midrule
\multicolumn{4}{l}{\textbf{GB}} & Hyena \\
\multicolumn{2}{l}{demo\_coding\_vs\_intergenomic\_seqs} & 20 & & 202 \\
\multicolumn{2}{l}{demo\_human\_or\_worm}                & 20 & & 202 \\
\multicolumn{2}{l}{human\_nontata\_promoters}            & 20 & & 253 \\
\multicolumn{2}{l}{human\_enhancers\_cohn}               & 20 & & 502 \\
\multicolumn{2}{l}{human\_enhancers\_ensembl}            & 20 & & 550 \\
\multicolumn{2}{l}{human\_ocr\_ensembl}                  & 20 & & 579 \\
\multicolumn{2}{l}{human\_ensembl\_regulatory}           & 20 & & 802 \\
\bottomrule
\end{tabular}
\end{adjustbox}
\end{table}

\subsection{Frozen probing benchmarks}

On BEND gene finding and DART-Eval Task 4 the pretrained model is used only as a
feature extractor. We follow each benchmark's released protocol without
modification: embeddings are computed in a single forward pass with all
pretrained weights frozen, and only the benchmark's own probe is trained on
them. For BEND this is a nine-class convolutional probe over per-nucleotide
embeddings, evaluated by MCC on the test split. For DART-Eval
Task 4, it is a convolutional regression probe trained on peak, non-peak, and IDR
regions of the K562 cell line and evaluated on held-out chromosomes by Spearman
correlation.

\FloatBarrier
\section{Full Pruning Sweeps}
\label{app:sweeps}

This section reports the full pruning sweeps for each pretraining setting,
including variants omitted from Section~\ref{sec:results}. The tables show
how different pruning thresholds affect data savings, pretraining cost,
and downstream performance. The full sweeps also reveal how far each method can be pushed before downstream performance deteriorates. For the encoder settings, SP remains effective over a wider range of DSI
before downstream performance deteriorates. On DB2--MSG, BPAS degrades
sharply beyond roughly 55\% DSI, whereas SP remains competitive with the
unpruned baseline at 77.77\% DSI
(Table~\ref{tab:full-sweep-db2-msg}). Across the encoder sweeps, the more
aggressive SP\_10\% setting marks the point where downstream performance
begins to drop substantially.

The Hyena sweep shows a different pattern. BPAS performs better at low DSI,
but the performance gap narrows as DSI increases. SP\_50\% achieves the best
downstream result among the SP variants and exceeds the unpruned baseline
while using roughly half of the baseline pretraining data. The sweep also
exposes a structural limit of BPAS. Because BPAS measures the change in
activation statistics between consecutive epochs, it requires two completed
epochs before pruning can begin. These first two full epochs cap its maximum
DSI at 92.86\%. SP can prune after the first epoch and reaches 96.42\% DSI
while remaining close to the unpruned baseline
(Table~\ref{tab:full-sweep-hyenadna-hr}).

\begin{table*}[t]
\renewcommand{\tabcolsep}{3.5pt}
\centering
\caption{Detailed pruning sweep for DB2 pretraining on MSG.}
\label{tab:full-sweep-db2-msg}
\begin{adjustbox}{max width=\textwidth}
\begin{tabular}{llcccccccc}
\toprule
Pruning
 & \multirow{2}{*}{Variant}
 & \multicolumn{2}{c}{Threshold}
 & DSI
 & GPU$\downarrow$
 & \multirow{2}{*}{Epch.}
 & \multirow{2}{*}{Steps}
 & \multicolumn{2}{c}{MCC$\uparrow$} \\
\cmidrule(lr){3-4}\cmidrule(lr){9-10}
Apprch. & & Low & High & (\%)$\uparrow$ & Hours & & & GUE & NT-18 \\
\midrule

\multirow{2}{*}{Baseline}
 & Unpruned & - & - & 0 & 1012.13 & 47 & 500k & \underline{71.58} & \textbf{69.79} \\
 & Full pruned & - & - & 100 & 0 & 0 & 0 & 60.57 & 47.12 \\

\cmidrule(lr){1-10}
\multirow{9}{*}{Related}
 & BPAS\_1 & $2\!\times\!10^{-6}$ & $2\!\times\!10^{-5}$ & 6.67 & 977.99 & 47 & 470k & 71.31 & 68.83 \\
\multirow{9}{*}{Work} & BPAS\_2 & $1\!\times\!10^{-5}$ & $1\!\times\!10^{-4}$ & 19.81 & 777.84 & 47 & 400k & 71.17 & \underline{69.59} \\
 & BPAS\_3 & $7\!\times\!10^{-5}$ & $7\!\times\!10^{-4}$ & 34.61 & 652.99 & 35 & 320k & 71.02 & 68.70 \\
 & BPAS\_4 & $1\!\times\!10^{-4}$ & $1\!\times\!10^{-3}$ & 36.25 & 838.74  & 38 & 310k & 71.03 & 68.59 \\
 & BPAS\_5 & $2\!\times\!10^{-4}$ & $2\!\times\!10^{-3}$ & 40.72 & 752.49  & 33 & 290k & 70.67 & 68.67 \\
 & BPAS\_6 & $3\!\times\!10^{-4}$ & $3\!\times\!10^{-3}$ & 45.36 & 689.09  & 30 & 270k & 70.19 & 68.34 \\
 & BPAS\_7 & $4\!\times\!10^{-4}$ & $4\!\times\!10^{-3}$ & 46.35 & 666.24 & 29 & 270k & 70.30 & 68.69 \\
 & BPAS\_8 & $1\!\times\!10^{-3}$ & $1\!\times\!10^{-2}$ & 54.69 & 473.71 & 25 & 220k & 70.12 & 68.16 \\
 & BPAS\_9 & $5\!\times\!10^{-3}$ & $5\!\times\!10^{-2}$ & 71.23 & 318.55 & 24 & 140k & 68.88 & 67.52 \\
 & BPAS\_10 & $1\!\times\!10^{-2}$ & $1\!\times\!10^{-1}$ & 89.91 & 100.61 & 9 & 50k & 66.30 & 63.34 \\

\cmidrule(lr){1-10}
\multirow{12}{*}{Ours}
 & SP\_100\% & 100 & 100 & 5.53 & 994.88 & 47 & 480k & 71.54 & 69.54 \\
 & SP\_90\% & 90 & 100 & 8.05 & 946.69 & 47 & 460k & 70.55 & 69.28 \\
 & SP\_80\% & 80 & 100 & 12.07 & 905.06 & 47 & 430k & 70.76 & 69.15 \\
 & SP\_70\% & 70 & 100 & 16.67 & 834.11 & 47 & 410k & 70.54 & 68.77 \\
 & SP\_60\% & 60 & 100 & 23.14 & 786.63 & 47 & 380k & \textbf{72.00} & 69.10 \\
 & SP\_50\% & 50 & 100 & 32.42 & 713.61 & 47 & 330k & 71.49 & 69.00 \\
 & SP\_40\% & 40 & 100 & 42.29 & 494.22 & 47 & 280k & 71.35 & 68.60 \\
 & SP\_30\% & 30 & 100 & 57.19 & 438.88 & 47 & 210k & 70.30 & 69.28 \\
 & SP\_20\% & 20 & 100 & 77.77 & 226.51 & 47 & 110k & 70.47 & 69.05 \\
 & SP\_10\% & 10 & 100 & 93.20 & 72.52 & 47 & 30k & 65.86 & 62.08 \\
 & SP\_5\% & 5 & 100 & 96.65 & 30.12 & 47 & 16k & 63.32 & 60.70 \\
 & SP\_1\% & 1 & 100 & 96.76 & 27.80 & 47 & 15k & 63.65 & 60.39 \\

\cmidrule(lr){1-10}
\multirow{2}{*}{Ablation}
 & RP\_20\% & - & - & 77.97 & 215.92 & 47 & 110k & 68.12 & 65.57 \\
 & LP\_0\% & 0 & 0 & 55.84 & 378.18 & 47 & 230k & 68.53 & 64.85 \\

\bottomrule
\end{tabular}
\end{adjustbox}
\end{table*}

\begin{table*}[t]
\renewcommand{\tabcolsep}{3.5pt}
\centering
\caption{Selected pruning configurations for DB2 pretraining on HRG.}
\label{tab:selected-sweep-db2-hr}
\begin{adjustbox}{max width=\textwidth}
\begin{tabular}{llcccccccc}
\toprule
Pruning
 & \multirow{2}{*}{Variant}
 & \multicolumn{2}{c}{Threshold}
 & DSI
 & GPU$\downarrow$
 & \multirow{2}{*}{Epch.}
 & \multirow{2}{*}{Steps}
 & \multicolumn{2}{c}{MCC$\uparrow$} \\
\cmidrule(lr){3-4}\cmidrule(lr){9-10}
Apprch. & & Low & High & (\%)$\uparrow$ & Hours & & & GUE & NT-18 \\
\midrule
\multirow{2}{*}{Baseline}
 & Unpruned & - & - & 0 & 47.16 & 32 & 30k & \textbf{66.61} & 64.14 \\
 & Full pruned & - & - & 100 & 0 & 0 & 0 & 43.56 & 47.27 \\

\cmidrule(lr){1-10}
\multirow{5}{*}{Related}
 & BPAS\_1 & $2\!\times\!10^{-6}$ & $2\!\times\!10^{-5}$ & 0.72 & 47.15 & 32 & 30k & 65.75 & 62.72 \\
\multirow{5}{*}{Work} & BPAS\_4 & $1\!\times\!10^{-4}$ & $1\!\times\!10^{-3}$ & 29.63 & 32.95 & 31 & 21k & 64.99 & 62.41 \\
 & BPAS\_5 & $2\!\times\!10^{-4}$ & $2\!\times\!10^{-3}$ &  40.40 & 23.57 & 29 & 18k & 64.37 & 62.83  \\
 & BPAS\_6 & $3\!\times\!10^{-4}$ & $3\!\times\!10^{-3}$ & 59.21 & 19.24 & 25 & 12k & 64.70 & 61.78 \\
 & BPAS\_8 & $1\!\times\!10^{-3}$ & $1\!\times\!10^{-2}$ & 91.03 & 4.31 & 8 & 3k & 62.42 & 61.82 \\
 & BPAS\_9 & $5\!\times\!10^{-3}$ & $5\!\times\!10^{-2}$ & 93.75 & 2.99 & 2 & 2k & 61.40 & 60.96 \\

\cmidrule(lr){1-10}
\multirow{6}{*}{Ours}
 & SP\_100\% & 100 & 100 & 2.88 & 46.78 & 32 & 29k & 65.20 & 63.80 \\
 & SP\_80\% & 80 & 100 & 5.93 & 45.04 & 32 & 28k & 65.42 & \textbf{65.70} \\
 & SP\_50\% & 50 & 100 & 18.12 & 39.16 & 32 & 25k & \underline{66.12} & 63.57 \\
 & SP\_20\% & 20 & 100 & 46.29 & 25.78 & 32 & 16k & 65.20 & 63.24 \\
 & SP\_10\% & 10 & 100 & 72.00 & 13.47 & 32 & 8k & 62.44 & 60.17 \\
 & SP\_5\% & 5 & 100 & 89.22 & 5.23 & 32 & 3k & 61.54 & 60.09 \\

\cmidrule(lr){1-10}
\multirow{4}{*}{Ablation}
& EL2N\_1 & 0 & 0.70 & 15.93 & 39.91 & 32 & 25k & 64.16 & \underline{64.70} \\
& EL2N\_2 & 0 & 0.80 & 25.98 & 34.99 & 32 & 22k & 65.32 & 64.61 \\
& EL2N\_3 & 0 & 0.90 & 37.65 & 32.57 & 32 & 19k & 65.75 & 64.21 \\
& EL2N\_4 & 0 & 0.95 & 44.96 & 27.83 & 32 & 16k & 64.98 & 63.03 \\

\bottomrule
\end{tabular}
\end{adjustbox}
\end{table*}

\begin{table*}[t]
\renewcommand{\tabcolsep}{3.5pt}
\centering
\caption{Detailed pruning sweep for NT-500M pretraining on HRG.}
\label{tab:full-sweep-nt-hrg}
\begin{adjustbox}{max width=\textwidth}
\begin{tabular}{llcccccccc}
\toprule
Pruning
 & \multirow{2}{*}{Variant}
 & \multicolumn{2}{c}{Threshold}
 & DSI
 & GPU$\downarrow$
 & \multirow{2}{*}{Epch.}
 & \multirow{2}{*}{Steps}
 & \multicolumn{2}{c}{MCC$\uparrow$} \\
\cmidrule(lr){3-4}\cmidrule(lr){9-10}
Apprch. & & Low & High & (\%)$\uparrow$ & Hours & & & GUE & NT-18 \\
\midrule

\multirow{2}{*}{Baseline}
 & Unpruned & - & - & 0 & 1137.60 & 99 & 50k & 66.98 & \textbf{74.23} \\
 & Full pruned & - & - & 100 & 0 & 0 & 0 & 53.00 & 63.74 \\

\cmidrule(lr){1-10}
\multirow{4}{*}{Related}
 & BPAS\_1 & $2\!\times\!10^{-6}$ & $2\!\times\!10^{-5}$ & 1.56 & 1087.51 & 99 & 49k & 66.37 & 73.75 \\
\multirow{4}{*}{Work} & BPAS\_2 & $1\!\times\!10^{-5}$ & $1\!\times\!10^{-4}$ & 7.10 & 1033.89 & 99 & 46k & 65.60 & 73.70 \\
 & BPAS\_4 & $1\!\times\!10^{-4}$ & $1\!\times\!10^{-3}$ & 40.59 & 658.58 & 99 & 29k & 65.61 & 73.90 \\
 & BPAS\_8 & $1\!\times\!10^{-3}$ & $1\!\times\!10^{-2}$ & 73.71 & 284.79 & 56 & 13k & 63.38 & 72.61 \\
& BPAS\_9 & $5\!\times\!10^{-3}$ & $5\!\times\!10^{-2}$ & 82.61 & 215.70 & 31 & 9k & 61.14 & 71.43 \\

\cmidrule(lr){1-10}
\multirow{5}{*}{Ours}
 & SP\_100\% & 100 & 100 & 4.78 & 1081.86 & 99 & 47k & 67.37 & 73.90 \\
 & SP\_80\% & 80 & 100 & 6.66 & 1063.21 & 99 & 46k & 67.09 & 73.88 \\
 & SP\_50\% & 50 & 100 & 10.31 & 969.95 & 99 & 44k & \textbf{68.19} & 74.01 \\
 & SP\_20\% & 20 & 100 & 37.94 & 668.83 & 99 & 30k & \underline{67.44} & \underline{74.12} \\
 & SP\_10\% & 10 & 100 & 65.41 & 379.00 & 99 & 17k & 64.46 & 73.18 \\

\bottomrule
\end{tabular}
\end{adjustbox}
\end{table*}

\begin{table*}[t]
\renewcommand{\tabcolsep}{3.5pt}
\centering
\caption{Detailed pruning sweep for Hyena pretraining on HRG.}
\label{tab:full-sweep-hyenadna-hr}
\begin{adjustbox}{max width=\textwidth}
\begin{tabular}{llccccccc}
\toprule
Pruning
 & \multirow{2}{*}{Variant}
 & \multicolumn{2}{c}{Threshold}
 & DSI
 & GPU$\downarrow$
 & \multirow{2}{*}{Epch.}
 & \multirow{2}{*}{Steps}
 & MCC$\uparrow$ \\
\cmidrule(lr){3-4}
Apprch. & & Low & High & (\%)$\uparrow$ & Hours & & & GB \\
\midrule
\multirow{2}{*}{Baseline}
 & Unpruned & - & - & 0 & 76.41 & 28 & 50k & 76.97 \\
 & Full pruned & - & - & 100 & 0 & 0 & 0 & 73.27 \\

\cmidrule(lr){1-9}
\multirow{7}{*}{Related}
 & BPAS\_1 & $2\!\times\!10^{-6}$ & $2\!\times\!10^{-5}$ & 0.001 & 75.20 & 28 & 49k & 76.65 \\
\multirow{7}{*}{Work} & BPAS\_4 & $1\!\times\!10^{-4}$ & $1\!\times\!10^{-3}$ & 2.67 & 73.14 & 28 & 48k & 77.18 \\
 & BPAS\_6 & $3\!\times\!10^{-4}$ & $3\!\times\!10^{-3}$ & 9.26 & 68.09 & 28 & 45k & \underline{77.31} \\
 & BPAS\_8 & $1\!\times\!10^{-3}$ & $1\!\times\!10^{-2}$ & 37.42 & 46.96 & 28 & 30k & 76.83 \\
 & BPAS\_9 & $5\!\times\!10^{-3}$ & $5\!\times\!10^{-2}$ & 63.52 & 27.54 & 23 & 18k & 77.19 \\
 & BPAS\_10 & $1\!\times\!10^{-2}$ & $1\!\times\!10^{-1}$ & 74.65 & 19.14 & 23 & 13k & \textbf{77.64} \\
 & BPAS\_11 & $1\!\times\!10^{-1}$ & $1\!\times\!10^{0}$ & 84.13 & 15.81 & 7 & 7.8k & 76.70\\
 & BPAS\_12 & $7\!\times\!10^{1}$ & $1\!\times\!10^{2}$ & 92.86 & 4.65 & 2 & 4k & 76.05\\

\cmidrule(lr){1-9}
\multirow{6}{*}{Ours}
 & SP\_100\% & 100 & 100 & 4.37 & 71.73 & 28 & 47k & 76.42 \\
 & SP\_80\% & 80 & 100 & 11.79 & 65.94 & 28 & 43k & 76.37 \\
 & SP\_50\% & 50 & 100 & 47.07 & 39.75 & 28 & 26k & \underline{77.31} \\
 & SP\_30\% & 30 & 100 & 96.38 & 2.8 & 28 & 1.7k & 75.37 \\
 & SP\_20\% & 20 & 100 & 96.42 & 2.8 & 28 & 1.7k & 76.11 \\
 & SP\_10\% & 10 & 100 & 96.43 & 2.78 & 28 & 1.7k & 75.28 \\
\cmidrule(lr){1-9}
\multirow{5}{*}{Ablation}
 & Perp\_1 & 1.00 & 1.13 & 6.92 & 71.20 & 28 & 47k & 75.85 \\
 & Perp\_2 & 1.00 & 2.19 & 20.52 & 51.77 & 28 & 39k & 76.47 \\

 & Perp\_3 & 1.00 & 3.00 & 45.86 & 36.52 & 28 & 27k & 76.70 \\
 & Perp\_4 & 1.00 & 12.00 & 96.42 & 2.19 & 28 & 1.7k & 75.36 \\
\cmidrule(lr){2-9}
 & RP\_20\% & - & - & 96.43 & 3.62 & 28 & 1.7k & 74.81  \\

\bottomrule
\end{tabular}
\end{adjustbox}
\end{table*}

\FloatBarrier
\section{Hyperparameters and computation}
\label{app:hparams}

Table~\ref{tab:variant-parameters} lists the variant-specific parameters
used across the pretraining settings. For BPAS, $(\delta_s,\delta_e)$
are the starting and ending values of its epoch-dependent
activation-change threshold. For InfoBatch, $(r,\delta)$ are the
drop probability and annealing fraction. SP, EL2N, Perp, and LP use
low and high thresholds on the pruning score range; RP has no
threshold parameters.

\begin{table*}[htbp]
\centering
\caption{Variant-specific parameters across pretraining settings.}
\label{tab:variant-parameters}

\renewcommand{\tabcolsep}{1pt}
\renewcommand{\arraystretch}{0.92}

\begin{adjustbox}{max width=\textwidth,center}
\begin{tabular}{lcc@{\hspace{12pt}}lcc@{\hspace{12pt}}lcc}
\toprule
\multicolumn{3}{c}{Related Work} &
\multicolumn{3}{c}{Ours} &
\multicolumn{3}{c}{Ablation} \\
\cmidrule(lr){1-3}
\cmidrule(lr){4-6}
\cmidrule(lr){7-9}
\multirow{2}{*}{Variant} & \multicolumn{2}{c}{Parameters} &
\multirow{2}{*}{Variant} & \multicolumn{2}{c}{Threshold} &
\multirow{2}{*}{Variant} & \multicolumn{2}{c}{Threshold} \\
\cmidrule(lr){2-3}
\cmidrule(lr){5-6}
\cmidrule(lr){8-9}
 & $\delta_s$/$r$ & $\delta_e$/$\delta$ &
 & $\tau_{\mathrm{low}}$ & $\tau_{\mathrm{high}}$ &
 & Low & High \\
\midrule

BPAS\_1
 & $2\!\times\!10^{-6}$ & $2\!\times\!10^{-5}$
 & SP\_100\% & 100 & 100
 & EL2N\_1 & 0 & 0.70 \\

BPAS\_2
 & $1\!\times\!10^{-5}$ & $1\!\times\!10^{-4}$
 & SP\_90\% & 90 & 100
 & EL2N\_2 & 0 & 0.80 \\

BPAS\_3
 & $7\!\times\!10^{-5}$ & $7\!\times\!10^{-4}$
 & SP\_80\% & 80 & 100
 & EL2N\_3 & 0 & 0.90 \\

BPAS\_4
 & $1\!\times\!10^{-4}$ & $1\!\times\!10^{-3}$
 & SP\_70\% & 70 & 100
 & EL2N\_4 & 0 & 0.95 \\

\cmidrule(lr){7-9}

BPAS\_5
 & $2\!\times\!10^{-4}$ & $2\!\times\!10^{-3}$
 & SP\_60\% & 60 & 100
 & Perp\_1 & 1.00 & 1.13 \\

BPAS\_6
 & $3\!\times\!10^{-4}$ & $3\!\times\!10^{-3}$
 & SP\_50\% & 50 & 100
 & Perp\_2 & 1.00 & 2.19 \\

BPAS\_7
 & $4\!\times\!10^{-4}$ & $4\!\times\!10^{-3}$
 & SP\_40\% & 40 & 100
 & Perp\_3 & 1.00 & 3.00 \\

BPAS\_8
 & $1\!\times\!10^{-3}$ & $1\!\times\!10^{-2}$
 & SP\_30\% & 30 & 100
 & Perp\_4 & 1.00 & 12.00 \\

\cmidrule(lr){7-9}

BPAS\_9
 & $5\!\times\!10^{-3}$ & $5\!\times\!10^{-2}$
 & SP\_20\% & 20 & 100
 & RP\_20\% & - & - \\

BPAS\_10
 & $1\!\times\!10^{-2}$ & $1\!\times\!10^{-1}$
 & SP\_10\% & 10 & 100
 & LP\_0\% & 0 & 0 \\

BPAS\_11
 & $1\!\times\!10^{-1}$ & $1\!\times\!10^{0}$
 & SP\_5\% & 5 & 100
 & & & \\

BPAS\_12
 & $7\!\times\!10^{1}$ & $1\!\times\!10^{2}$
 & SP\_1\% & 1 & 100
 & & & \\

\cmidrule(lr){1-3}

Info\_1
 & 0.5 & 0.875
 & & &
 & & & \\

Info\_2
 & 0.7 & 0.90
 & & &
 & & & \\

\bottomrule
\end{tabular}
\end{adjustbox}
\end{table*}

\par
\Needspace{20\baselineskip}
\begin{wraptable}{R}{0.68\textwidth}
\caption{Fine-tuning seed variance. Each entry is the mean $\pm$ standard deviation over three fine-tuning seeds using a single pretrained checkpoint.}
\label{tab:ft-variance}
\centering
\renewcommand{\tabcolsep}{2pt}
\begin{adjustbox}{max width=\linewidth}
\begin{tabular}{lllccc}
\toprule
\multirow{2}{*}{Corpus}
 & \multirow{2}{*}{Model}
 & \multirow{2}{*}{Variant}
 & \multicolumn{3}{c}{MCC$\uparrow$} \\
\cmidrule(lr){4-6}
 & & & GUE & NT-18 & GB \\
\midrule
\multirow{2}{*}{MSG} & \multirow{2}{*}{DB2} & Unpruned & $70.80 \pm 0.52$ & $68.98 \pm 0.52$ & ---\\
 & & SP\_20\% & $70.22 \pm 0.29$ & $68.85 \pm 0.18$ & ---\\
\midrule
\multirow{2}{*}{HRG} & \multirow{2}{*}{DB2} & Unpruned & $66.29 \pm 0.57$ & $63.53 \pm 0.60$ & ---\\
 & & SP\_20\% & $64.98 \pm 0.36$ & $62.99 \pm 0.11$ & ---\\
\midrule
\multirow{2}{*}{HRG} & \multirow{2}{*}{NT-500M} & Unpruned & $66.46 \pm 0.52$ & $74.10 \pm 0.13$ & ---\\
 & & SP\_20\% & $66.98 \pm 0.40$ & $73.67 \pm 0.14$ & ---\\
\midrule
\multirow{2}{*}{HRG} & \multirow{2}{*}{Hyena} & Unpruned & --- & --- & $76.51 \pm 0.50$\\
 & & SP\_20\% & --- & --- & $75.34 \pm 0.31$\\
\bottomrule
\end{tabular}
\end{adjustbox}
\end{wraptable}

\FloatBarrier
\section{Run-to-Run Variance}
\label{app:variance}

We separately measure variation introduced by downstream fine-tuning and by pretraining. This allows us to assess whether the observed performance and data-savings differences are robust to random seeds.

\subsection{Fine-Tuning Seed Variance}
\label{app:ft-variance}
We fine-tune each checkpoint with three different random seeds while
keeping the downstream settings fixed. Table~\ref{tab:ft-variance}
reports the mean and standard deviation of downstream MCC for the
unpruned baseline and SP\_20\%. This isolates fine-tuning variation
without changing the pretrained checkpoint.

\subsection{Pretraining Seed Variance}
\label{app:pt-variance}

We pretrain SP\_20\% with three different random seeds while keeping
the pretraining settings fixed, then fine-tune each checkpoint once.
Table~\ref{tab:pt-variance} reports the mean and standard deviation
of DSI and downstream MCC. We focus these repeats on SP\_20\% because
its lower computation cost makes repeated pretraining substantially less
expensive than for the unpruned baselines.

\begin{table*}[t]
\caption{Pretraining seed variance for SP\_20\%. Each entry is the mean $\pm$ standard deviation over three independent pretraining runs, each fine-tuned once.}
\label{tab:pt-variance}
\centering
\renewcommand{\tabcolsep}{3.5pt}
\begin{adjustbox}{max width=\textwidth}
\begin{tabular}{lllcccc}
\toprule
\multirow{2}{*}{Corpus}
 & \multirow{2}{*}{Model}
 & \multirow{2}{*}{Variant}
 & DSI
 & \multicolumn{3}{c}{MCC$\uparrow$} \\
\cmidrule(lr){5-7}
 & & & (\%) & GUE & NT-18 & GB \\
\midrule
MSG & DB2 & \multirow{4}{*}{SP\_20\%} & $77.61 \pm 0.15$ & $70.18 \pm 0.28$ & $68.60 \pm 0.17$ & --- \\
HRG & DB2 & & $46.18 \pm 0.13$ & $64.34 \pm 0.50$ & $62.82 \pm 0.59$ & --- \\
HRG & NT-500M & & $37.85 \pm 0.08$ & $67.15 \pm 0.25$ & $73.77 \pm 0.38$ & ---\\
HRG & Hyena & & $96.42 \pm 0.00$ & --- & --- & $75.77 \pm 0.34$ \\
\bottomrule
\end{tabular}
\end{adjustbox}
\end{table*}

\section{DSI-Targeted Sequence Pruning}
\label{app:dsi}

Let $N$ be the number of sequences in the unpruned corpus and $I$ the scheduled
number of epochs, indexed by $i\in\{1,\ldots,I\}$. Let
$\mathcal{A}_i\subseteq\{1,\ldots,N\}$ be the active set entering epoch $i$,
with $\mathcal{A}_1=\{1,\ldots,N\}$, and let $p$ index an individual sequence.
Finally, let $I_s\leq I$ denote the last executed epoch. The running DSI through
epoch $i$ is
\begin{equation}
\mathrm{DSI}_i
=
1-\frac{\sum_{j=1}^{i} |\mathcal{A}_j|}{i\,N},
\label{eq:dsi_2}
\end{equation}
the average fraction of the corpus left untouched through epoch $i$. Reading
this equation backwards turns a target DSI curve into a pretraining budget. At the
boundary after epoch $i$, the target $\mathrm{DSI}_{i+1}$ determines how many
sequences may remain in the next epoch. We remove the sequences with the highest
valid $\mathrm{Acc}_p^{(i)}$ from $\mathcal{A}_i$ until $\mathcal{A}_{i+1}$
meets that budget.

The default curve comes from measurement rather than choice. Across our
SP runs, the retained fraction in epoch $i$ is well described by
the geometric decay
\begin{equation}
\frac{|\mathcal{A}_i|}{N}=\rho^{\,i-1}.
\end{equation}
For a target final DSI $D$, we fit $\rho$ so that the final DSI under the equation in section~\ref{paragraph:eqs} equals $D$. The other three curves change when pruning
occurs while keeping the same final target. Replacing the exponent $i-1$ with
$(i-1)^{1/g}$ gives
\begin{equation}
\frac{|\mathcal{A}_i|}{N}
=
\rho^{(i-1)^{1/g}},
\end{equation}
where $g=4$ gives the front-loaded curve and $g=0.35$ gives the back-loaded
curve. The linear curve instead uses
\begin{equation}
\frac{|\mathcal{A}_i|}{N}
=
1-2m(i-1),
\qquad
m=\frac{D}{I-1}.
\end{equation}
For $D=0.5$ and $I=47$, the resulting values are $\rho=0.9657$ for the
default curve, $\rho=0.7086$ for the front-loaded curve, $\rho=0.9999$ for the
back-loaded curve, and $m=0.0109$ for the linear curve. Under the linear
schedule, $|\mathcal{A}_{47}|=0$, so the last executed epoch is $I_s=46$. The
running DSI at that point is $\mathrm{DSI}_{46}=0.489$, but the final DSI
reported is $0.5$ because the scheduled horizon
remains $I=47$.

During epoch $i$, we accumulate
$M_{p,\mathrm{correct}}^{(i)}$ and $M_{p,\mathrm{total}}^{(i)}$, producing
$\mathrm{Acc}_p^{(i)}$ as defined in the Equation in section~\ref{sec:method}. Sequences that are not
processed because they fall outside a complete batch, or for which
$M_{p,\mathrm{total}}^{(i)}=0$, are ranked last and are therefore removed only
when the budget cannot be met from the scored sequences alone, as happens when a
schedule targets an empty active set. The learning-rate schedule remains tied to
the original total step count and is never rescaled, so the pruned and unpruned
runs remain comparable step for step.

\end{document}